\documentclass{article}

\PassOptionsToPackage{numbers, compress}{natbib}
\usepackage[preprint]{neurips_2026/neurips_2026}

\usepackage[utf8]{inputenc} % allow utf-8 input
\usepackage[T1]{fontenc}    % use 8-bit T1 fonts
\usepackage{hyperref}       % hyperlinks
\usepackage{url}            % simple URL typesetting
\usepackage{booktabs}       % professional-quality tables
\usepackage{amsfonts}       % blackboard math symbols
\usepackage{nicefrac}       % compact symbols for 1/2, etc.
\usepackage{microtype}      % microtypography
\usepackage{xcolor}         % colors
\usepackage{amsmath}
\usepackage{graphicx}
\usepackage{pifont}% http://ctan.org/pkg/pifont
\newcommand{\cmark}{\ding{51}}%
\newcommand{\xmark}{\ding{55}}%
\usepackage{svg}

\title{Improving the Accuracy of Amortized Model Comparison with Self-Consistency}

\author{%
  Šimon~Kucharský \\
  Department of Statistics \\
  TU Dortmund University \\
  Dortmund, Germany \\
  \texttt{simon.kucharsky@tu-dortmund.de} \\
  \And
  Aayush~Mishra \\
  Department of Statistics \\
  TU Dortmund University \\
  Dortmund, Germany \\
  \texttt{aayush.mishra@tu-dortmund.de} \\
  \And
  Daniel~Habermann \\
  Department of Statistics \\
  TU Dortmund University \\
  Dortmund, Germany \\
  \texttt{daniel.habermann@tu-dortmund.de} \\
  \And
  Stefan T.~Radev \\
  Department of Cognitive Science \\
  Rensselaer Polytechnic Institute \\
  Troy, NY, USA \\
  \texttt{radevs@rpi.edu} \\
  \And
  Paul-Christian~Bürkner \\
  Department of Statistics \\
  TU Dortmund University \\
  Dortmund, Germany \\
  \texttt{paul.buerkner@tu-dortmund.de} \\
}

\begin{document}

\maketitle

\begin{abstract}
    Amortized Bayesian model comparison (BMC) enables fast probabilistic ranking of models via simulation-based training of neural surrogates. However, the accuracy of neural surrogates deteriorates when simulation models are misspecified; the very case where model comparison is most needed. We evaluate four different amortized BMC methods. We supplement traditional simulation-based training of these methods with a \emph{self-consistency} (SC) loss on unlabeled real data to improve BMC estimates under distribution shifts. Using one artificial and two real-world case studies, we compare amortized BMC estimators with and without SC against analytic or bridge sampling benchmarks. In the \emph{closed-world} case (data is generated by one of the candidate models), BMC estimators using classifiers work acceptably well even without SC training. However, these methods also benefit the least from SC training. In the \emph{open-world} scenario (all models misspecified), SC training strongly improves BMC estimators when having access to analytic likelihoods, or when surrogate likelihoods are locally accurate near the true parameter posterior, even for severely misspecified models. We conclude with practical recommendations for amortized BMC and suggestions for future research.
\end{abstract}

\section{Introduction}

Bayesian model comparison (BMC) is a formal method to rank competing probabilistic models according to their compatibility with observed data \citep{mackay2003information}. By combining prior model probabilities with evidence provided by the data, BMC yields posterior model probabilities that quantify plausibility of each model in consideration. However, BMC requires solving an integral that is often analytically or numerically intractable for nontrivial models \citep{lotfi2022bayesian}.

Modern machine learning approaches can circumvent intractable integrals by training neural surrogates to approximate model-implied quantities, such as posterior model probabilities \citep{radev2021amortized}, Bayes factors \citep{jeffrey2024evidence}, or marginal likelihoods \citep{radev2023jana}. Typically, these networks are trained on data simulated from the statistical models under consideration, thus falling under the umbrella of simulation-based inference \citep{cranmer2020frontier}. Once trained, the networks can rapidly approximate quantities that would otherwise demand costly re-computation for each new dataset. In this way, we trade initial training cost for substantial gains in efficiency during inference. This property is called \emph{amortization}, giving the name amortized Bayesian inference (ABI), and in the context of model comparison, amortized BMC.

Despite their efficiency gains, neural networks are notoriously susceptible to out-of-distribution \citep[OOD;][]{yang2024generalized} data. In standard ABI, neural networks are trained on data simulated from the statistical models. If some of the candidate models are misspecified relative to the true data-generative process, the real observed data might be OOD, and inferences based on the neural networks can be arbitrarily wrong \citep{schmitt2023detecting, frazier2024statistical}. When comparing different models, there is an inherent assumption that some models fit the data better than others, and some (or even all) models might be misspecified. This creates a \textit{conundrum}: neural estimates cannot be trusted for models that fit the data poorly, yet proper BMC requires faithful estimates even for poorly fitting models. 
The recently proposed self-consistency (SC) loss family \citep{schmitt_leveraging_2024,ivanova_data-efficient_2024,mishra_robust_2025} is a promising candidate to help solve the issue of OOD-related biases. SC does not require model simulations and can be even used with unlabeled real data. This allows the neural networks to generalize to data inaccessible during pure simulation-based training. 

\begin{figure}[t]
    \centering
    \includegraphics[width=\linewidth]{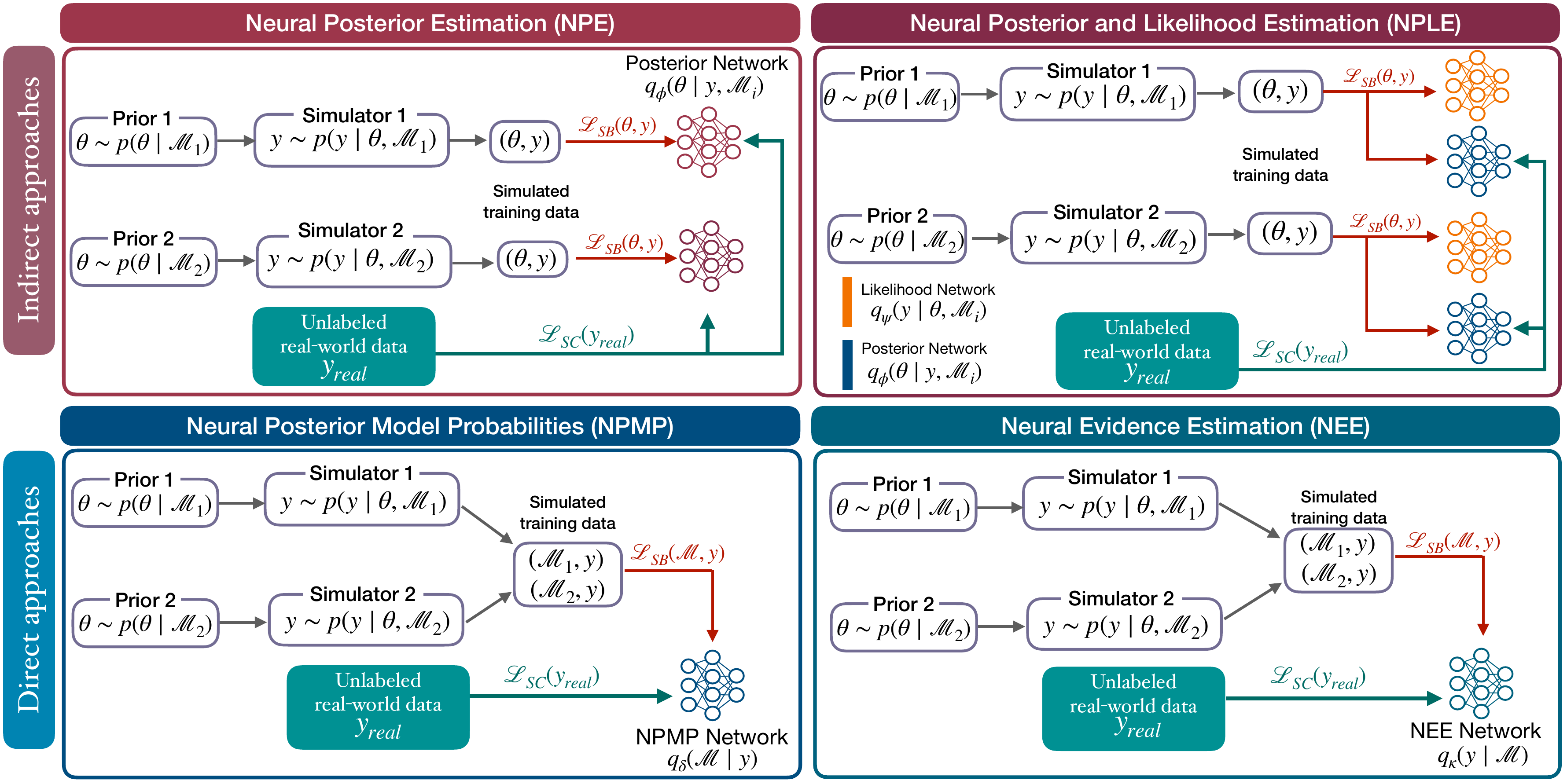}
    \caption{Training stage overview of the four amortized BMC approaches. \textit{Top}: indirect approaches that infer per-model parameter posteriors and derive model evidence from them. \textit{Bottom}: direct approaches that learn either posterior model probabilities or evidences. All methods incorporate unsupervised self-consistency losses to improve accuracy under model misspecification.
    }
    \label{fig:train_overview}
\end{figure}

In this work, we study the accuracy of amortized BMC methods and ask whether SC training improves their estimates especially when simulation-based training alone would be inadequate.
We extend the notion of self-consistency losses to direct approaches for model comparison, enabling SC training not only for posterior-based estimators but also for methods that directly approximate marginal likelihoods or posterior model probabilities.
We evaluate four approaches to amortized BMC shown in \autoref{fig:train_overview} across different scenarios, comparing their estimates against gold-standard baselines.
\section{Bayesian model comparison}
Given a set of $K$ candidate models $\mathcal{M} = \{M_1,\dots,M_K\}$ with model-specific random parameters $\theta_1, \dots, \theta_K$, the parameter posterior for each model conditioned on data $y$ follows Bayes' rule 
\begin{equation}
\label{eq:parameter_posterior}
    p(\theta_k \mid y, M_k)
 = \frac{p(\theta_k \mid M_k) \; p(y \mid \theta_k, M_k)}{p(y \mid M_k)}.
 \end{equation}
The denominator is the \emph{marginal likelihood} (i.e., model evidence),
\begin{equation}
\label{eq:marginal_likelihood}
p(y\mid M_k)=\int p(\theta_k\mid M_k) \; p(y\mid \theta_k,M_k) d\theta_k.
\end{equation}
Besides normalizing the posterior, the marginal likelihood plays a crucial role in BMC. A ratio of two marginal likelihoods $\text{BF}_{ij} := p(y \mid M_i) / p(y \mid M_j)$ captures the relative prior predictive advantage of model $M_i$ over $M_j$.
Applying Bayes' rule to the marginal likelihood gives us the posterior model probability (PMP):
\begin{equation}
\label{eq:bayes_theorem_pmp}
p(M_k\mid y)=\frac{p(M_k) \; p(y\mid M_k)}{\sum_{j=1}^K p(M_j) \; p(y\mid M_j)},
\end{equation}
which ranks models according to their marginal likelihoods weighted by their prior model probabilities $p(M_k)$.
The BF is sometimes reported over PMPs, as it is independent of the prior model probabilities \citep{kass1995bayes,jeffreys1948theory}. A set of marginal likelihoods is sufficient to compute PMPs and BFs, but not \textit{vice versa}. %However, PMP estimates can always be converted into BF estimates, but not \textit{vice versa}; a set of marginal likelihood estimates is sufficient for both quantities.

Unfortunately, for non-trivial models, the marginal likelihood (integral in Eq.~\ref{eq:marginal_likelihood}) is often intractable. Popular methods for approximating the marginal likelihood \citep[e.g., bridge sampling,][]{gronau2020bridgesampling} require sampling many draws from the posterior $p(\theta_k\mid y,M_k)$ using MCMC \citep{llorente2023marginal,knuth2015bayesian}, which is itself computationally demanding. This motivates the use of amortized methods for model comparison.

\section{Amortized Bayesian inference and model comparison}

ABI approximates a conditional distribution $p(z \mid y)$ of latent variable $z$ conditioned on observables $y$ with a generative network $q_\phi(z \mid y)$ trained on $N$ simulated pairs \smash{$(z^{(n)},y^{(n)})\sim p(z,y)$}. The network typically minimizes a simulation-based loss:
\begin{equation}
\label{eq:abi_unified}
\mathcal{L}_{\textrm{SB}} = \frac{1}{N}\sum_{n=1}^N S \big(q_\phi(\cdot \mid y^{(n)}),\, z^{(n)} \big).
\end{equation}
Provided that $S$ is a strictly proper scoring rule \citep[e.g., the log score $-\log q_\phi(z^{(n)} \mid y^{(n)})$;][]{gneiting2007strictly} and $q_\phi$ is sufficiently expressive \citep{draxler2024universality}, minimizing the loss encourages $q_\phi(z\mid y)\to p(z\mid y)$ as $N\to\infty$.

After training, the distribution $p(z \mid y^\text{obs})$ for any observed data set $y^\text{obs}$ can be approximated by sampling draws $z \sim q_\psi(z \mid y^\text{obs})$. This step takes only a little amount of time; inference on real data is instantaneous compared to training.
Because Eq.~\ref{eq:abi_unified} requires only simulation from, not evaluation of, $p(z,y)$, ABI is a subset of simulation-based inference \citep[SBI;][]{cranmer2020frontier}.

%For reasons that will become apparent below, we use network architectures which provide robust and fast density evaluation of $q_\phi$, such as coupling flows \citep{rezende2015variational,dinh2016density,durkan2019neural}. 

We consider four different approaches to amortized BMC, which differ in the quantities they approximate and the data used to train them. Broadly, the methods fall into two categories:  
(i) approaches that estimate the \emph{parameter-posterior} and derive the marginal likelihood from it, and  
(ii) approaches that estimate \emph{model-level quantities} directly. 

\subsection{Indirect approaches} 

The first two methods can indirectly estimate the marginal likelihood of a single model by targeting either the posterior or both the posterior and the likelihood \citep{radev2023jana}. They rely on rearranging the log of Bayes' rule for the parameter posterior (Eq.~\ref{eq:parameter_posterior}), which lets us express the log marginal likelihood in terms of prior, likelihood, and posterior,
\begin{equation}
\label{eq:marginal_likelihood_rearranged}
    \log p(y \mid M_k) = \log p(\theta_k \mid M_k) + \log p(y \mid \theta_k, M_k) - \log p(\theta_k \mid y, M_k),
\end{equation}
which forms the basis for the two estimators described below. For both,
simulation-based training relies on data generated from the specific model,
\begin{equation}
\label{eq:model_specific_simulation}
    \theta_k^{(n)} \sim p(\theta_k \mid M_k), \quad
    y^{(n)} \sim p(y \mid \theta_k^{(n)}, M_k).
\end{equation}

\paragraph{Neural posterior estimation (NPE).} NPE approximates the parameter posterior $p(\theta_k \mid y, M_k)$ with a generative network $q_\phi(\theta_k \mid y, M_k)$ trained on simulations from Eq.~\ref{eq:model_specific_simulation} by minimizing
\begin{equation}
    \mathcal{L}_{\textrm{SB}} = -\frac{1}{N} \sum_{n=1}^N \log q_\phi(\theta_k^{(n)} \mid y^{(n)}, M_k).
\end{equation}
It estimates the marginal likelihood by substituting $q_\phi$ into Eq.~\ref{eq:marginal_likelihood_rearranged}. However, the estimate based on a single $\theta_k$ may be unreliable due to approximation error caused by $q_\phi \neq p$. To mitigate this, we compute a Monte Carlo estimate over posterior samples \smash{$\tilde{\theta}_k^{(s)} \sim q_\phi(\theta \mid y, M_k)$}:
\begin{equation}
\label{eq:log_ml_npe_monte_carlo}
    \log p(y \mid M_k) \approx \frac{1}{S} \sum_{s=1}^S 
    \big[\log p(\tilde{\theta}_k^{(s)} \mid M_k) + \log p(y \mid \tilde{\theta}_k^{(s)}, M_k) - \log q_\phi(\tilde{\theta}_k^{(s)} \mid y, M_k)\big].
\end{equation}

\paragraph{Neural posterior and likelihood estimation (NPLE).} When the likelihood $p(y \mid \theta_k, M_k)$ is unavailable, we can approximate both posterior and likelihood with networks $q_\phi(\theta_k \mid y, M_k)$ and $q_\psi(y \mid \theta_k, M_k)$, trained on simulations from Eq.~\ref{eq:model_specific_simulation}. Alternatively, the likelihood network could be trained on data simulated with a different proposal distribution for the parameters:
\begin{equation}
    \theta_k^{(n)} \sim p^{*}(\theta_k), \quad y^{(n)} \sim p(y\mid\theta_k^{(n)}, M_k).
\end{equation}
As long as the observables $y$ are generated from the correct data distribution, the likelihood network will approximate the correct likelihood (potentially in a different region of the parameter space defined by the proposal $p^{*}$).
The marginal likelihood is estimated by replacing the posterior and likelihood in Eq.~\ref{eq:marginal_likelihood_rearranged} with $q_\phi$ and $q_\psi$, respectively, and averaging over posterior samples as in Eq.~\ref{eq:log_ml_npe_monte_carlo}.

\subsection{Direct approaches}

The remaining two methods bypass parameter-posterior estimation and instead learn model-level quantities (posterior model probability or marginal likelihood) directly from simulations of the entire set of the compared models,
\begin{equation}
\label{eq:model_ensemble_simulation}
    \begin{aligned}
        M_k^{(n)} \sim p(M_k), \quad
        \theta_k^{(n)} \sim p(\theta_k \mid M_k^{(n)}), \quad
        y^{(n)} \sim p(y \mid \theta_k^{(n)}, M_k^{(n)}).
    \end{aligned}
\end{equation}

\paragraph{Neural posterior model probabilities (NPMP).} NPMP estimates $K$ posterior model probabilities (PMPs) conditioned on data $p(M_k \mid y)$. PMPs can be estimated via a classifier $q_\delta$ (e.g., an MLP with a softmax output layer) trained with a cross-entropy loss \citep{radev2021amortized,elsemuller2024deep},
\begin{equation}
\begin{aligned}
    \mathcal{L}_{\textrm{SB}} = -\frac{1}{N} \sum_{n=1}^N\log q_\delta\big(M_k^{(n)} \mid y^{(n)}\big).
\end{aligned}
\end{equation}

\paragraph{Neural evidence estimation (NEE).} Alternatively, the marginal likelihood (evidence) $p(y \mid M_k)$ can be approximated with a generative network $q_\kappa$ directly from the simulations in Eq~\ref{eq:model_ensemble_simulation}. We call this approach \emph{neural evidence estimation} (NEE). Simulation-based training is set up to minimize the negative log likelihood with one-hot encoded model index as the conditioning variable,
\begin{equation}
\begin{aligned}
    \mathcal{L}_{\textrm{SB}} =  -\frac{1}{N} \sum_{n=1}^N\log q_\kappa\big(y^{(n)} \mid M_k^{(n)} \big).
\end{aligned}
\end{equation}

Both NPMP and NEE are trained \textit{jointly} on simulations from all $K$ models. In contrast, indirect methods (NPE, NPLE) require a separate training loop for each candidate model.

\subsection{Amortized BMC and OOD data}
\label{sec:ood}

The notion of OOD data differs between the methods discussed above. NPE and NPLE are trained on simulations from a single model, and are therefore accurate within the generative scope of that model; data generated by another model may be OOD. In contrast, NPMP and NEE are trained on simulations drawn from the entire model set (Eq.~\ref{eq:model_ensemble_simulation}), so any data within the generative scope of at least one of the models is effectively in-distribution. In a \textit{closed-world} scenario (i.e., the true data-generating mechanism is included among the candidate models), NPMP and NEE are therefore expected to produce accurate estimates, whereas NPE and NPLE may be less reliable, depending on the similarity between the compared models. In an \textit{open-world} scenario, where the true data-generating process lies outside the model class, all four methods may become inaccurate to varying degrees.

\subsection{Self-consistency}

Self-consistency (SC) training is a promising candidate to make neural estimators more accurate outside of the generative scope of the training model(s) \citep{schmitt_leveraging_2024,mishra_robust_2025}. SC expresses the marginal likelihood in terms of prior, likelihood, and posterior (Eq.~\ref{eq:marginal_likelihood_rearranged}) and leverages the fact that for a given dataset $y$, the marginal likelihood is always constant, regardless of which parameter values $\theta_k \in \Theta_k$ are plugged in the equation. Estimates of the marginal likelihood with surrogate densities should therefore also be invariant to different parameter values. We define the SC loss as the variance of the log marginal likelihood estimate, averaged over $M$ unlabeled datasets $y^{(1)}, \dots, y^{M}$,
\begin{equation}
\begin{aligned}
\label{eq:sc_loss}
    \mathcal{L}_{\textrm{SC}} = \frac{1}{M} & \sum_{m=1}^M \mathrm{Var}_{\tilde{\theta}_k\sim p_c(\theta_k)} \big[\log p(\tilde{\theta}_k \mid M_k) + \log p\big(y^{(m)} \mid \tilde{\theta}_k, M_k\big) -  \log q_\phi\big(\tilde{\theta}_k \mid y^{(m)}, M_k\big) \big],
\end{aligned}
\end{equation}
with a proposal distribution $p_c$ set to the current approximate posterior $q_\phi(\theta_k \mid y^{(m)}, M_k)$ \citep{mishra_robust_2025}. When the likelihood is unknown, it is replaced with its surrogate as well. We estimate Eq.~\ref{eq:sc_loss} by constructing a Monte Carlo estimate from the approximate posterior samples obtained by the posterior network.

For direct approaches, the same logic can be applied on the level of the model implied quantities, where rearranging Eq.~\ref{eq:bayes_theorem_pmp} yields,
\begin{equation}
    p(y) = \sum_{k=1}^K p(M_k) p(y \mid M_k) =  \frac{p(M_k)\,p(y\mid M_k)}{p(M_k\mid y)} \quad \forall k = 1,\dots,K.
\end{equation}
This implies that the marginal likelihood of the entire model set $p(y)$ can be expressed in terms of the prior model probabilities, marginal likelihood of the individual models, and the posterior model probabilities. Building on this, we introduce a novel self-consistency loss, obtained by substituting the marginal likelihood with NEE estimates and the posterior model probabilities with NPMP estimates,
\begin{equation}
\begin{aligned}
    \mathcal{L}_{\textrm{SC}} = \frac{1}{M} & \sum_{m=1}^M \mathrm{Var}_{\tilde{M}_k\sim p_c(M_k)} \big[\log p(\tilde{M}_k) + \log q_\kappa\big(y^{(m)} \mid \tilde{M}_k\big) -  \log q_\delta\big(\tilde{M}_k \mid y^{(m)}\big) \big]
\end{aligned}
\end{equation}
Here, $p_c$ can be a simple proposal distribution, such as the prior model probabilities (our experiments use a discrete uniform distribution).

We train all networks using a weighted combination of the simulation-based and self-consistency losses, $\mathcal{L}_{\text{total}} = \mathcal{L}_{\textrm{SB}} + \lambda_{\text{SC}} \mathcal{L}_\textrm{SC}$.

\paragraph{SC breaks the correspondence between model misspecification and OOD.} The main advantage of SC training is that the data sets used for calculating the SC loss need not be labeled. We only require observables $y^{(m)}$ that need not be generated from any of the models; even empirical data without ground truth labels can be used. SC can therefore extend the training distribution beyond the generative scope of the model(s), without changing the underlying model. As a consequence, data that would be OOD under simulation-based training can become effectively in-distribution after SC training, even for severely misspecified models. This is especially important for amortized BMC where the data could be in the tails of the marginal likelihoods (see Section~\ref{sec:ood}).

\paragraph{SC without likelihoods.} When both posterior and (marginal) likelihood are approximate, the SC loss is not strictly proper \citep{mishra_robust_2025}: minimizing the loss does not guarantee that the networks converge to the correct solution, as the posterior and likelihood surrogates can drift towards a wrong solution albeit in a mutually self-consistent way. To tackle this issue, we train only one of the networks (that is, either the likelihood or the posterior) with SC, while the other is trained only with the corresponding simulation-based loss. Table~\ref{tab:methods} provides an overview of the methods.

\begin{table*}[t]
    \centering
    \caption{Overview of amortized model comparison methods paired with SC training. Columns indicate whether the method requires likelihood access, supports parameter inference, is trained per model or jointly across models, and which component is optimized with the SC loss. \textsuperscript{\dag} Strictly proper SC loss.}
    \begin{tabular}{lcccl}
    \toprule
       Method  &  Requires likelihood & Parameter inference & Single-model & SC training \\
    \midrule
       NPE  &  \cmark & \cmark & \cmark & posterior\textsuperscript{\dag} \\
       NPLE &  \xmark & \cmark & \cmark & posterior \\
       NPMP & \xmark & \xmark & \xmark & posterior \\
       NEE & \xmark & \xmark & \xmark & likelihood \\
    \bottomrule
    \end{tabular}
    %\caption{Overview of the methods for amortized model comparison using SC training. \emph{Requires likelihood:} Does the method require likelihood to work. \emph{Parameter inference:} Can the trained networks be used for parameter inference in addition for model comparison. \emph{Single-model:} Are the networks trained on simulations from a single model, or are they trained on simulations from all models under consideration. \emph{SC training:} Which network is being trained with the SC loss. \textsuperscript{\dag} Strictly proper SC loss.}
    %\caption{Overview of amortized model comparison methods with SC training. Columns indicate whether the method requires likelihood access, supports parameter inference, is trained per model or jointly across models, and which component is optimized with the SC loss. \textsuperscript{\dag} Strictly proper SC loss.}
    \label{tab:methods}
\end{table*}

\section{Related work}

\citet{radev2021amortized} studied amortized BMC with classifiers, an approach applied across multiple domains \citep[e.g.,][]{karchev2023simsims, sokratous2023ask, schumacher2025validation}. An extension for multilevel models was brought by \citet{elsemuller2024deep}. \citet{schroder2024simultaneous} enabled joint parameter inference and model comparison. \citet{jeffrey2024evidence} surveyed different losses for estimating Bayes factors and PMPs. \citet{elsemuller2024sensitivity} used deep ensembles for OOD detection. The JANA framework allowed approximating the marginal likelihood using joint training of posterior and likelihood networks \citep{radev2023jana}. \citet{spurio2023bayesian} proposed post-processing posterior samples with a learned harmonic mean estimator, and \citet{srinivasan2024bayesian} used normalizing flows to estimate marginal likelihoods from posterior samples. \citet{schmitt_leveraging_2024} showed that the SC loss improves marginal likelihood estimation for well-specified models. No mentioned work studied or attempted to rectify the ``extrapolation bias'' \citep{frazier2024statistical} under OOD conditions, except the ensemble detection approach of \citet{elsemuller2024sensitivity}. We use the semi-supervised SC loss of \citet{mishra_robust_2025} to assess its potential for reliable BMC under model misspecification.

\section{Case studies}

In the following three case studies, we compare the four amortized BMC approaches with and without SC training. The case studies cover common scenarios that arise when comparing statistical models; see  \autoref{app:case-studies} for a conceptual comparison of these cases.

\subsection{Multivariate Gaussian}
\label{case-mv-gaussian}

\begin{figure}[t]
    \centering
    \includegraphics[width=\linewidth]{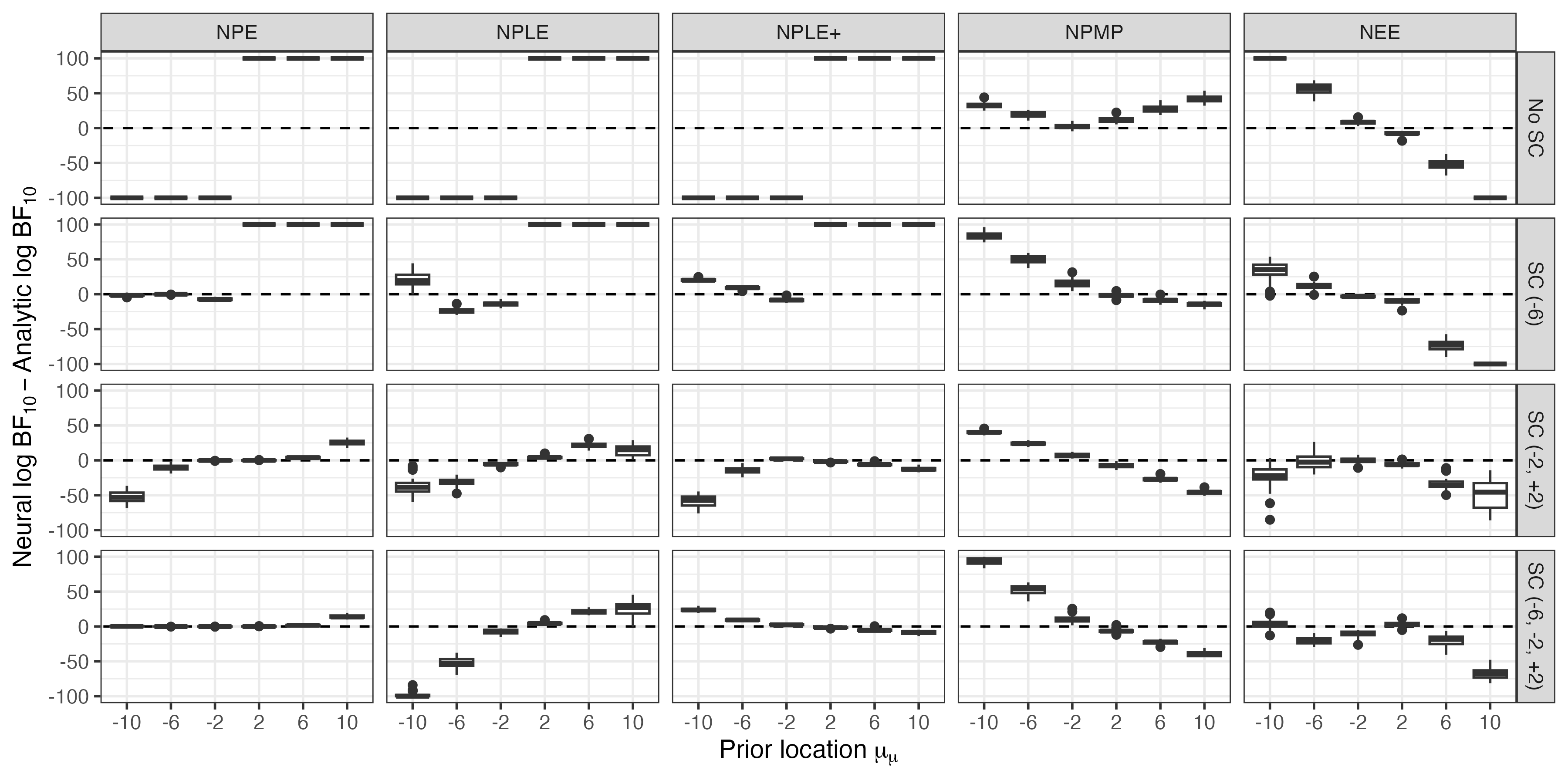}
    \caption{\textbf{Multivatiate Gaussian case study.} The error of the approximate log Bayes factors, for the four amortized methods (plus NPLE+, see Section~\ref{case-mv-gaussian}) under four different SC training conditions. Data along the \emph{y}-axis is squished at -100 and 100.}
    \label{fig:gaussian_log_bf_diff}
\end{figure}

In our first case study, we present an example of two models whose overlap is relatively small. We demonstrate how SC training influences Bayes factor estimation, and the effects of choosing different data sets for SC training.

We use a toy example of $D$-dimensional Gaussian location models,
\begin{equation}
    \begin{aligned}
    \boldsymbol\mu_y \sim \mathcal{N}(\boldsymbol\mu_\mu,\, \mathbf{I}_D), \quad
    \mathbf{y} \sim \mathcal{N}(\boldsymbol\mu_y,\, \mathbf{I}_D).
    \end{aligned}
\end{equation}
We compare two models specified with a prior shift: \(\mathcal{M}_0: \boldsymbol\mu_\mu = -2\), \(\mathcal{M}_1: \boldsymbol\mu_\mu = 2\), and we set $D=10$. All neural methods were trained with simulation-based loss, combined with the SC loss in four conditions. These four conditions vary what data was used for the SC loss; \emph{No SC}: No SC training, \emph{SC (-6)}: data generated from a model with $\boldsymbol\mu_\mu = -6$, representing an ``open-world'' scenario where both models are misspecified to the available data, \emph{SC (-2, +2)}: data generated from the two compared models, \emph{SC (-6, -2, +2)}: combined data from the previous two conditions.

Accuracy of the results was evaluated on data sets with varying location $\boldsymbol\mu_\mu$ between -10 and 10, but with the same likelihood, testing the networks' robustness against prior misspecification. We also consider an additional method (NPLE+) where the likelihood network is pre-trained on simulations using prior $\boldsymbol\mu_y \sim \text{Uniform}(-10, 10)$ independently drawn for each dimension. The likelihood network was subsequently frozen during training of the posterior network. By training the likelihood network to data generated under a broader range of parameter values, we expect the likelihood to become more robust to prior misspecification, and as an effect, improve the results of SC training. See \autoref{app:gaussian} for more details about the experimental setup and additional results.

\paragraph{Results.} In  \autoref{fig:gaussian_log_bf}, the log Bayes factors in favor of \(\mathcal{M}_1\) over \(\mathcal{M}_0\) from all neural methods are compared against the analytic solution. \textbf{NPE}: Without SC training, model comparison results are extremely inaccurate. This is because any dataset is OOD for at least one of the two models. With SC (-6), accuracy is improved for data ranging from $\boldsymbol\mu_\mu = -10$ to $\boldsymbol\mu_\mu = -2$ as both approximate posteriors learn to generalize towards negative values. SC (-2,+2) is effective as well particularly at $\boldsymbol\mu_\mu = \pm2$, but the benefits are visible for other data sets as well as the two approximate posteriors learn to generalize in both directions, though with an increasing error towards the extreme ends. SC (-6,-2,+2) provides the most accurate results that generalize beyond the the training data in both directions (with higher error remaining on the extreme positive end). \textbf{NPLE}: SC does have positive effects on BF estimation, especially in regions where the likelihood network is well trained due to simulation-based training ($\boldsymbol\mu_\mu = \pm2$). Outside of those regions the estimates are overall better than without SC training, but still with a substantial error. \textbf{NPLE+}: Compared to NPLE, the results are closer to NPE, demonstrating that training the likelihood on a wider range of parameter values can improve robustness of NPLE against prior misspecification. \textbf{NPMP}: Without SC the estimates are the most accurate of all methods, to some extent even outside of the generative scope of the two models, even though with decreasing accuracy towards the extreme ends. SC training pulls NPMP towards NEE which has a larger error. As a result, the estimates become more biased. \textbf{NEE}: Without SC training the estimates are severely biased for data outside of the generative scope of the two models. SC training increases accuracy (NEE is pulled towards NPMP which is more accurate) particularly in regions close to the SC data, but retains larger error on the extreme ends.

\subsection{Air passenger traffic forecasting}

\begin{figure}[t]
    \centering
    \includegraphics[width=\linewidth]{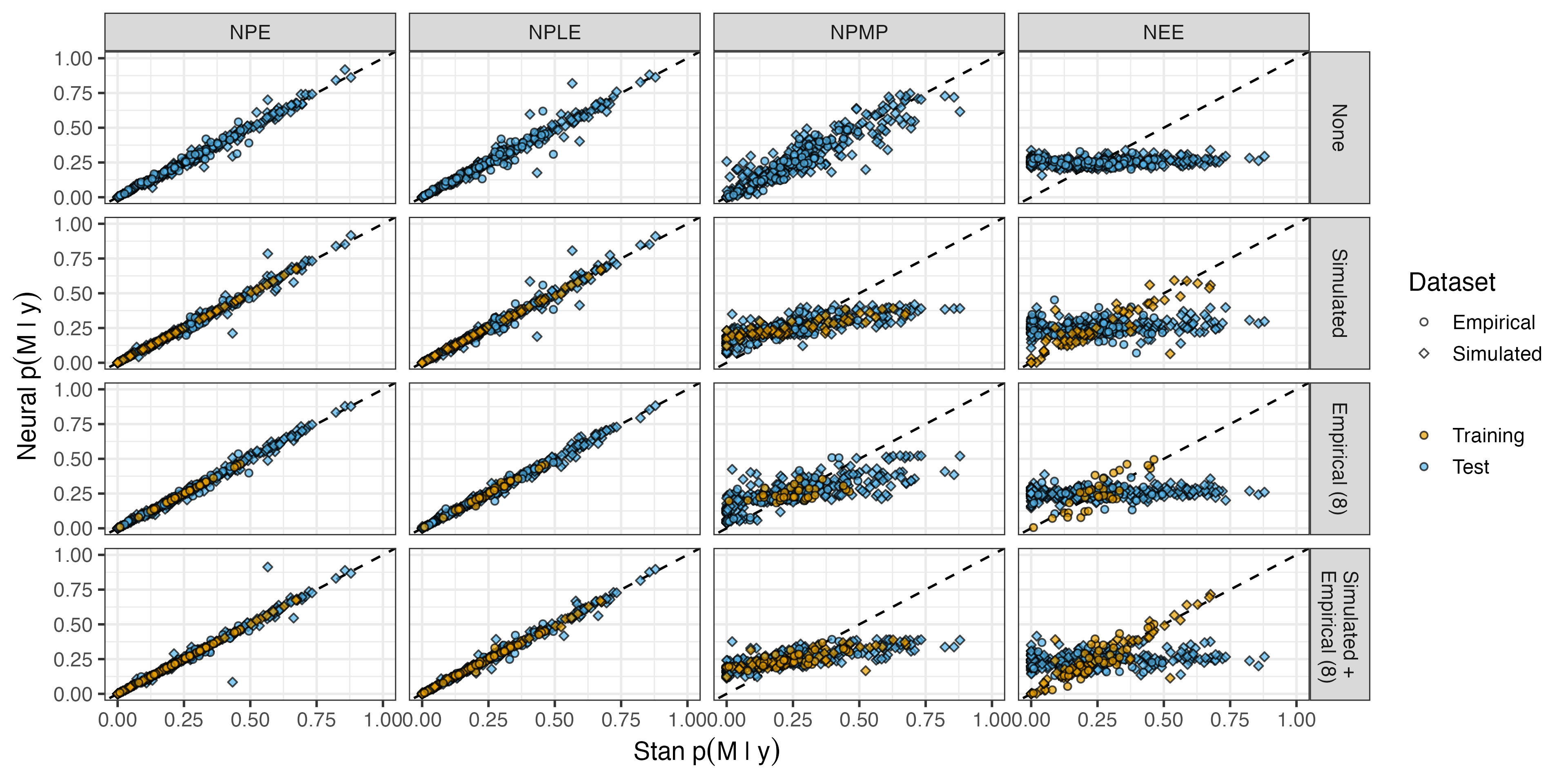}
    \caption{\textbf{Air traffic case study.} Posterior model probability estimates of the four methods compared to the gold standard obtained from 40,000 samples using Stan. Each model is evaluated on empirical data and simulations from the four models.}
    \label{fig:air_traffic_pmp_subset}
\end{figure}

Second, we investigate a scenario of four similar models, where the empirical data is close to in-distribution for all four models.

We analyze the European air passenger traffic dataset \citep{Eurostat2022a, Eurostat2022b, Eurostat2022c}, which has been used previously in the context of ABI \citep{habermann2024amortized,mishra_robust_2025}. This dataset contains annual counts of air passengers traveling from 15 European countries to the United States between 2004 and 2019, alongside each country's GDP and household debt indicators. We fit four statistical models that differ in their predictor sets for forecasting annual passenger departures: the ``Full'' model includes the previous year's departures (lag), household debt, and GDP, while the three reduced models each omit one of these predictors. All neural methods were trained with simulation-based loss, combined with the SC loss in different conditions. These conditions vary what data was used for computing the SC loss: \emph{None}: No SC loss was computed, \emph{Simulated}: Four data sets simulated from each model (16 data sets in total), \emph{Empirical (8)}: Empirical data from eight countries (data sets from the remaining seven countries are used as hold-out, test data), \emph{Simulated + Empirical (8)}: Combined data sets from the previous two conditions. Detailed information about the case study and results from additional conditions are included in \autoref{app:air_traffic}.

\begin{figure}[t]
    \centering
    \includegraphics[width=\linewidth]{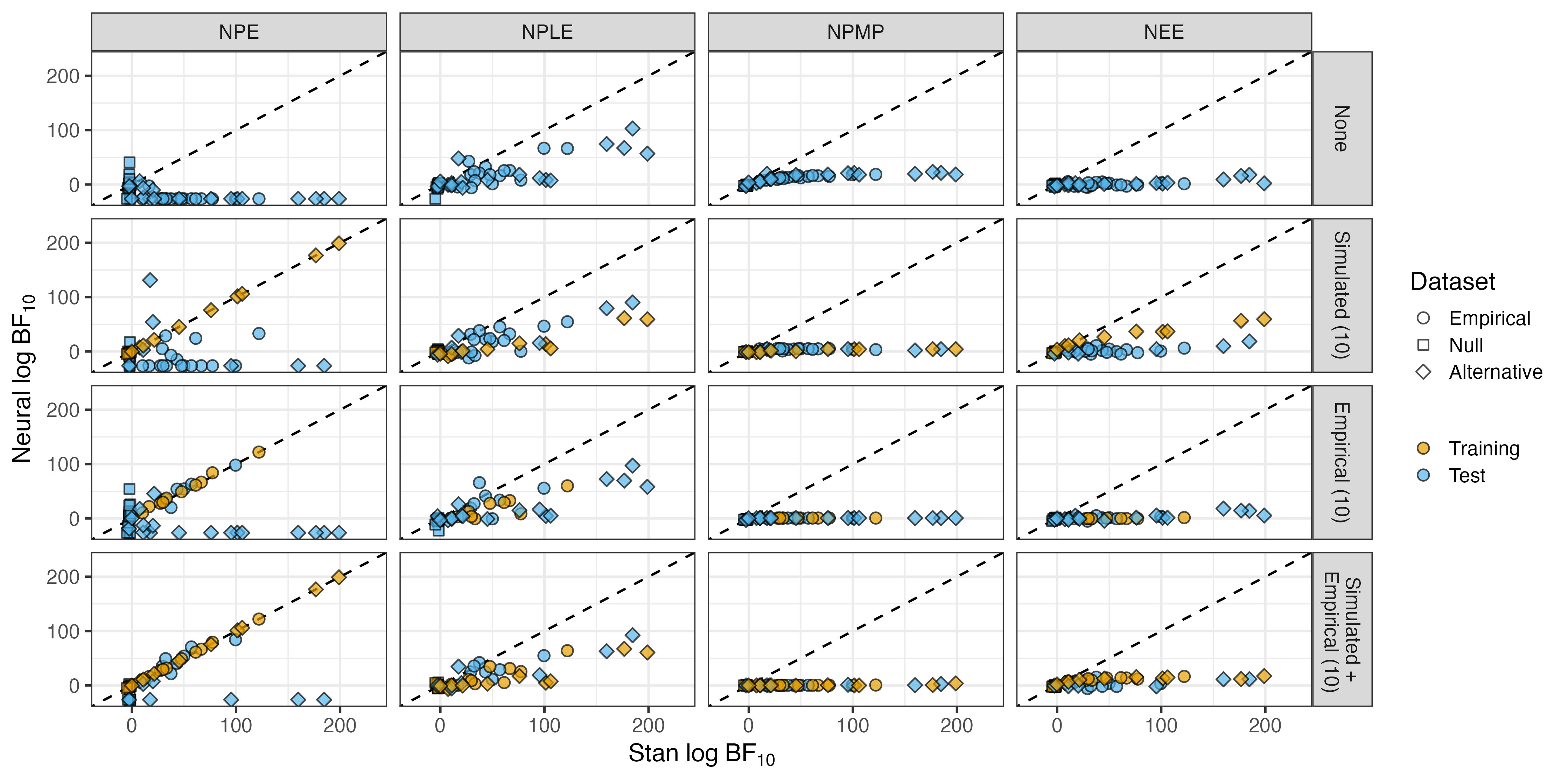}
    \caption{\textbf{Racing diffusion case study.} The log BF in favor of the alternative model for the four methods under four different SC training conditions compared to the gold standard obtained from 40,000 samples using Stan, evaluated on empirical datasets and datasets simulated from the two compared models.}
    \label{fig:diffusion_log_bf_subset}
\end{figure}

\paragraph{Results.} The neural posterior model probabilities are compared with the gold standard,  computed using bridge sampling \citep{gronau2020bridgesampling} with Stan \citep{stan, rstan}, and shown on \autoref{fig:air_traffic_pmp_subset}. \textbf{NPE} and \textbf{NPLE}: Estimates are pretty accurate even without SC training, and SC training improves the estimates even further especially for the data sets that were included in the SC training set. \textbf{NPMP} is relatively accurate without SC as well, albeit with a larger error than NPE and NPLE. SC training pulls NPMP towards NEE and increases its bias. \textbf{NEE}: Accuracy without SC training is poor, indicating insufficient learning from model simulations alone. SC training pulls NEE towards NPMP and improves its estimates. However, the network overfits to the SC training set.

\subsection{Racing diffusion model of decision making}

Lastly, we investigate a common scenario of two nested models. The ``null'' model assumes the nullity of an effect of a predictor and is a special case of the ``alternative'' model. In this case, Bayes factors are generally modest when the data come from the null model, but quickly diverge for data generated from the alternative model (or data for which both models are misspecified).

We analyze empirical data from the lexical decision task of \citet{wagenmakers2008diffusion} (Experiment 1), in which 17 participants classified letter strings as words or non-words, instructed to respond as fast as possible or accurate as possible. We model the choices and response times with the racing diffusion model \citep{tillman2020sequential}; the decision process is represented as noisy evidence accumulation: two evidence accumulators race toward a decision threshold $\alpha$. Under the null model, $\alpha$ is constrained to be constant across conditions (instructions); under the alternative model, it is allowed to vary. Similarly to the previous case study, we ran SC training in different conditions, depending on whether simulated or empirical data was used for training. Additional details about the case study are provided in \autoref{app:diffusion}. 

\paragraph{Results.} \autoref{fig:diffusion_log_bf_subset} shows the neural estimates of the log Bayes factor in favor of the alternative model, compared to the gold standard using bridge sampling \citep{gronau2020bridgesampling} with Stan \citep{stan, rstan}, evaluated on the 17 empirical datasets, simulations from the null model, and simulations from the alternative model. Note that for the empirical data and the data generated from the alternative model, the evidence in favor of the alternative model is so extreme that its PMP approaches 1. We also compare results on the PMP scale ( \autoref{fig:diffusion_pmp} and \ref{fig:diffusion_pmp_diff}). \textbf{NPE}: Evidence estimates are the most accurate when using SC training, but the network struggles to generalize reliably outside of the training SC set. \textbf{NPLE}: Results are more noisy than for NPE and benefits of SC training are not clearly visible. \textbf{NPMP}: The network is able to distinguish between the alternative and null model (by assigning PMP of $\approx 1$ to data from the alternative model or empirical data) without SC, but is unable to estimate Bayes factors correctly. This leads to generally higher error in assigning posterior model probabilities when data is generated from the null model. Training with SC worsens the estimates. \textbf{NEE}: Estimates are generally incorrect, SC improves the estimates to be more in line with NPMP.

\section{Discussion \& Limitations}

We investigated four methods for amortized model comparison and studied whether their generalization to OOD data can be improved by training with a self-consistency (SC) loss. 

NPE benefits the most from SC training. When the test data is close to the data used for SC training, NPE with SC can yield very accurate results, even when the models are misspecified. This is not a surprising result, since NPE has the advantage of access to the analytic likelihood. These findings suggest that, whenever available, amortized methods should leverage SC training in addition to the traditional simulation-based training. On the other hand, NPE+SC is limited to models that are possible to fit with likelihood-based methods as well (e.g., MCMC with bridge sampling), and not on models with intractable likelihood densities. Nevertheless, NPE might still be the preferred option for superior speed during inference.

When both posterior and likelihood are approximate (NPLE), the SC loss is no longer strictly proper and its effect becomes less reliable; in some settings it appears to help, in others it has little impact. Our assertion is that for NPLE to benefit from SC training, the likelihood surrogate must be trained well in the vicinity of the true parameter posterior. This conclusion is supported by the additional results from the Gaussian case study, where pretraining the likelihood on a wider range of parameters yielded comparable results to that of NPE. However, this fix required oracle knowledge on the parameter range in which to train. Without this knowledge, accurate results with NPLE are an obvious challenge, since the true parameter posterior is unknown. Despite these limitations, NPLE remains a promising approach when likelihoods are not available given a better alignment between the likelihood surrogate and the true posterior. We leave a detailed exploration of such approaches to future work.

NPMP proves itself adequate at least in two situations. First, in a closed-world setting, where the empirical data is known to be produced from one of the models under consideration, NPMP tends to yield relatively good estimates even without SC training. Second, when the evidence for one model is so extreme that its posterior probability is $\approx 1$, NPMP will correctly point to the correct model even if its estimate of the Bayes factors is biased. As for NEE, its results are the least flattering; though inaccurate without SC training, SC training aligns its performance to NPMP. Given that NPMP is typically less computationally expensive, there is little reason to use NEE.

We also note a tendency to overfit to the data sets used in SC training. This highlights the importance of adopting best practices to mitigate overfitting when applying SC, such as regularization, train-validation-test data splitting, and so forth.

\section{Conclusion}

Taken together, our findings lead to several practical recommendations. For practitioners interested in scalable and trustworthy Bayesian model comparison, we recommend using NPE finetuned with SC as the default amortized strategy when likelihood is available. When likelihoods are unavailable, NPMP without SC training remains a steady choice in closed-world scenarios. NPLE trained with SC is a promising candidate, but requires more careful approach and potentially alternative training methods.
Our results support the view that supplementing simulation-based training with training based on empirical data might be a powerful tool to address model misspecification in amortized methods \citep{mishra_robust_2025}. However, limitations of the methods without access to analytic likelihoods leave relevant gaps for further improvements.

\newpage
\bibliographystyle{plainnat}
{
\small
\bibliography{bibliography}

@article{cranmer2020frontier,
  title={The frontier of simulation-based inference},
  author={Cranmer, Kyle and Brehmer, Johann and Louppe, Gilles},
  journal={Proceedings of the National Academy of Sciences},
  volume={117},
  number={48},
  pages={30055--30062},
  year={2020},
  publisher={National Academy of Sciences}
}

@book{jeffreys1948theory,
  title={The theory of probability},
  author={Jeffreys, Harold},
  year={1948},
  publisher={Oxford University Press, Oxford},
}

@misc{mishra_robust_2025,
	title = {Robust {Amortized} {Bayesian} {Inference} with {Self}-{Consistency} {Losses} on {Unlabeled} {Data}},
	url = {http://arxiv.org/abs/2501.13483},
	doi = {10.48550/arXiv.2501.13483},
	urldate = {2025-07-09},
	publisher = {arXiv},
	author = {Mishra, Aayush and Habermann, Daniel and Schmitt, Marvin and Radev, Stefan T. and Bürkner, Paul-Christian},
	month = may,
	year = {2025},
	note = {arXiv:2501.13483 [stat]},
}

@inproceedings{lotfi2022bayesian,
  title={Bayesian model selection, the marginal likelihood, and generalization},
  author={Lotfi, Sanae and Izmailov, Pavel and Benton, Gregory and Goldblum, Micah and Wilson, Andrew Gordon},
  booktitle={International Conference on Machine Learning},
  pages={14223--14247},
  year={2022},
  organization={PMLR}
}

@book{mackay2003information,
  title={Information theory, inference and learning algorithms},
  author={MacKay, David JC},
  year={2003},
  publisher={Cambridge university press}
}

@misc{schmitt_leveraging_2024,
	title = {Leveraging {Self}-{Consistency} for {Data}-{Efficient} {Amortized} {Bayesian} {Inference}},
	url = {http://arxiv.org/abs/2310.04395},
	doi = {10.48550/arXiv.2310.04395},
	urldate = {2025-08-11},
	publisher = {arXiv},
	author = {Schmitt, Marvin and Ivanova, Desi R. and Habermann, Daniel and Köthe, Ullrich and Bürkner, Paul-Christian and Radev, Stefan T.},
	month = jul,
	year = {2024},
	note = {arXiv:2310.04395 [cs]},
}

@inproceedings{radev2023jana,
  title={{JANA: Jointly amortized neural approximation of complex Bayesian models}},
  author={Radev, Stefan T and Schmitt, Marvin and Pratz, Valentin and Picchini, Umberto and K{\"o}the, Ullrich and B{\"u}rkner, Paul-Christian},
  booktitle={Uncertainty in Artificial Intelligence},
  pages={1695--1706},
  year={2023},
  organization={PMLR}
}

@article{jeffrey2024evidence,
  title={{Evidence Networks: Simple losses for fast, amortized, neural Bayesian model comparison}},
  author={Jeffrey, Niall and Wandelt, Benjamin D},
  journal={Machine Learning: Science and Technology},
  volume={5},
  number={1},
  pages={015008},
  year={2024},
  publisher={IOP Publishing}
}

@article{frazier2024statistical,
  title={The statistical accuracy of neural posterior and likelihood estimation},
  author={Frazier, David T and Kelly, Ryan and Drovandi, Christopher and Warne, David J},
  journal={arXiv preprint arXiv:2411.12068},
  year={2024}
}

@article{yang2024generalized,
  title={Generalized out-of-distribution detection: {A} survey},
  author={Yang, Jingkang and Zhou, Kaiyang and Li, Yixuan and Liu, Ziwei},
  journal={International Journal of Computer Vision},
  volume={132},
  number={12},
  pages={5635--5662},
  year={2024},
  publisher={Springer}
}

@inproceedings{schmitt2023detecting,
  title={{Detecting model misspecification in amortized Bayesian inference with neural networks}},
  author={Schmitt, Marvin and B{\"u}rkner, Paul-Christian and K{\"o}the, Ullrich and Radev, Stefan T},
  booktitle={Dagm german conference on pattern recognition},
  pages={541--557},
  year={2023},
  organization={Springer}
}

@article{spurio2023bayesian,
  title={Bayesian model comparison for simulation-based inference},
  author={Spurio Mancini, A and Docherty, MM and Price, MA and McEwen, JD},
  journal={RAS Techniques and Instruments},
  volume={2},
  number={1},
  pages={710--722},
  year={2023},
  publisher={Oxford University Press}
}

@article{sokratous2023ask,
  title={How to ask twenty questions and win: {M}achine learning tools for assessing preferences from small samples of willingness-to-pay prices},
  author={Sokratous, Konstantina and Fitch, Anderson K and Kvam, Peter D},
  journal={Journal of choice modelling},
  volume={48},
  pages={100418},
  year={2023},
  publisher={Elsevier}
}

@article{srinivasan2024bayesian,
  title={Bayesian evidence estimation from posterior samples with normalizing flows},
  author={Srinivasan, Rahul and Crisostomi, Marco and Trotta, Roberto and Barausse, Enrico and Breschi, Matteo},
  journal={Physical Review D},
  volume={110},
  number={12},
  pages={123007},
  year={2024},
  publisher={APS}
}

@article{elsemuller2024sensitivity,
  year={2024},
  title={{Sensitivity-Aware Amortized Bayesian Inference}},
  author={Elsem{\"u}ller, Lasse and Olischl{\"a}ger, Hans and Schmitt, Marvin and B{\"u}rkner, Paul-Christian and Koethe, Ullrich and Radev, Stefan T},
  journal={Transactions on Machine Learning Research}
}

@article{draxler2024universality,
  title={On the universality of volume-preserving and coupling-based normalizing flows},
  author={Draxler, Felix and Wahl, Stefan and Schn{\"o}rr, Christoph and K{\"o}the, Ullrich},
  journal={arXiv preprint arXiv:2402.06578},
  year={2024}
}

@article{elsemuller2024deep,
  title={A deep learning method for comparing {B}ayesian hierarchical models.},
  author={Elsem{\"u}ller, Lasse and Schnuerch, Martin and B{\"u}rkner, Paul-Christian and Radev, Stefan T},
  journal={Psychological Methods},
  year={2024},
  publisher={American Psychological Association}
}

@inproceedings{schroder2024simultaneous,
  title={Simultaneous identification of models and parameters of scientific simulators},
  author={Schr{\"o}der, Cornelius and Macke, Jakob H},
  booktitle={Proceedings of the 41st International Conference on Machine Learning},
  pages={43895--43927},
  year={2024}
}

@article{schumacher2025validation,
  title={Validation and comparison of non-stationary cognitive models: {A} diffusion model application},
  author={Schumacher, Lukas and Schnuerch, Martin and Voss, Andreas and Radev, Stefan T},
  journal={Computational Brain \& Behavior},
  volume={8},
  number={2},
  pages={191--210},
  year={2025},
  publisher={Springer}
}

@article{karchev2023simsims,
  title={{SimSIMS: Simulation-based Supernova Ia Model Selection with thousands of latent variables}},
  author={Karchev, Konstantin and Trotta, Roberto and Weniger, Christoph},
  journal={arXiv preprint arXiv:2311.15650},
  year={2023}
}

@article{radev2021amortized,
  title={Amortized bayesian model comparison with evidential deep learning},
  author={Radev, Stefan T and D’Alessandro, Marco and Mertens, Ulf K and Voss, Andreas and Koethe, Ullrich and Buerkner, Paul-Christian},
  journal={IEEE Transactions on Neural Networks and Learning Systems},
  volume={34},
  number={8},
  pages={4903--4917},
  year={2021},
  publisher={IEEE}
}

@inproceedings{ivanova_data-efficient_2024,
	title = {Data-{Efficient} {Variational} {Mutual} {Information} {Estimation} via {Bayesian} {Self}-{Consistency}},
	url = {https://openreview.net/forum?id=QfiyElaO1f&noteId=aRvehpmMkK},
	language = {en},
	urldate = {2025-08-11},
	author = {Ivanova, Desi R. and Schmitt, Marvin and Radev, Stefan T.},
	month = oct,
	year = {2024},
    booktitle = {NeurIPS BDU Workshop 2024},
}

@Misc{Eurostat2022a,
  author    = {{Eurostat}},
  title     = {International extra-EU air passenger transport by reporting country and partner world regions and countries, {doi:10.2908/avia\_paexcc}},
  year      = {2022},
  doi       = {10.2908/AVIA_PAEXCC},
  publisher = {Eurostat},
}

@Misc{Eurostat2022b,
  author    = {{Eurostat}},
  title     = {{Household debt, consolidated including Non-profit institutions serving households - \% of GDP}, {doi:10.2908/TIPSD22}},
  year      = {2022},
  doi       = {10.2908/TIPSPD22},
  publisher = {Eurostat},
}

@Misc{Eurostat2022c,
  author    = {{Eurostat}},
  title     = {Real GDP per capita, {doi:10.2908/SDG\_08\_10}},
  year      = {2022},
  doi       = {10.2908/SDG_08_10},
  publisher = {Eurostat},
}

@article{habermann2024amortized,
  title={Amortized bayesian multilevel models},
  author={Habermann, Daniel and Schmitt, Marvin and K{\"u}hmichel, Lars and Bulling, Andreas and Radev, Stefan T and B{\"u}rkner, Paul-Christian},
  journal={arXiv preprint arXiv:2408.13230},
  year={2024}
}

@article{gronau2020bridgesampling,
  title={bridgesampling: An {R} package for estimating normalizing constants},
  author={Gronau, Quentin F and Singmann, Henrik and Wagenmakers, Eric-Jan},
  journal={Journal of Statistical Software},
  volume={92},
  pages={1--29},
  year={2020}
}

@article{wagenmakers2008diffusion,
  title={A diffusion model account of criterion shifts in the lexical decision task},
  author={Wagenmakers, Eric-Jan and Ratcliff, Roger and Gomez, Pablo and McKoon, Gail},
  journal={Journal of memory and language},
  volume={58},
  number={1},
  pages={140--159},
  year={2008},
  publisher={Elsevier}
}

@article{tillman2020sequential,
  title={Sequential sampling models without random between-trial variability: {T}he racing diffusion model of speeded decision making},
  author={Tillman, Gabriel and Van Zandt, Trish and Logan, Gordon D},
  journal={Psychonomic Bulletin \& Review},
  volume={27},
  number={5},
  pages={911--936},
  year={2020},
  publisher={Springer}
}

@Misc{rstan,
 title = {{RStan}: the {R} interface to {Stan}},
  author = {{Stan Development Team}},
  note = {R package version 2.32.7},
  year = {2025},
  url = {https://mc-stan.org/},
}

@Misc{stan,
 title = {{Stan Reference Manual}},
  author = {{Stan Development Team}},
  note = {version 2.32.2},
  year = {2025},
  url = {https://mc-stan.org/},
}

@article{kass1995bayes,
  title={Bayes factors},
  author={Kass, Robert E and Raftery, Adrian E},
  journal={Journal of the american statistical association},
  volume={90},
  number={430},
  pages={773--795},
  year={1995},
  publisher={Taylor \& Francis}
}

@article{knuth2015bayesian,
  title={Bayesian evidence and model selection},
  author={Knuth, Kevin H and Habeck, Michael and Malakar, Nabin K and Mubeen, Asim M and Placek, Ben},
  journal={Digital Signal Processing},
  volume={47},
  pages={50--67},
  year={2015},
  publisher={Elsevier}
}

@article{llorente2023marginal,
  title={Marginal likelihood computation for model selection and hypothesis testing: an extensive review},
  author={Llorente, Fernando and Martino, Luca and Delgado, David and Lopez-Santiago, Javier},
  journal={SIAM review},
  volume={65},
  number={1},
  pages={3--58},
  year={2023},
  publisher={SIAM}
}

@article{gneiting2007strictly,
  title={Strictly proper scoring rules, prediction, and estimation},
  author={Gneiting, Tilmann and Raftery, Adrian E},
  journal={Journal of the American statistical Association},
  volume={102},
  number={477},
  pages={359--378},
  year={2007},
  publisher={Taylor \& Francis}
}

@article{bayesflow_2023_software,
  title = {{BayesFlow}: Amortized {B}ayesian workflows with neural networks},
  author = {Radev, Stefan T. and Schmitt, Marvin and Schumacher, Lukas and Elsemüller, Lasse and Pratz, Valentin and Schälte, Yannik and Köthe, Ullrich and Bürkner, Paul-Christian},
  journal = {Journal of Open Source Software},
  volume = {8},
  number = {89},
  pages = {5702},
  year = {2023}
}

@article{bayesflow_2020_original,
  title = {{BayesFlow}: Learning complex stochastic models with invertible neural networks},
  author = {Radev, Stefan T. and Mertens, Ulf K. and Voss, Andreas and Ardizzone, Lynton and K{\"o}the, Ullrich},
  journal = {IEEE transactions on neural networks and learning systems},
  volume = {33},
  number = {4},
  pages = {1452--1466},
  year = {2020}
}

@article{dinh2016density,
  title={Density estimation using real nvp},
  author={Dinh, Laurent and Sohl-Dickstein, Jascha and Bengio, Samy},
  journal={arXiv preprint arXiv:1605.08803},
  year={2016}
}

@article{kuhmichel2026bayesflow,
  title={BayesFlow 2.0: Multi-Backend Amortized Bayesian Inference in Python},
  author={K{\"u}hmichel, Lars and Huang, Jerry M and Pratz, Valentin and Arruda, Jonas and Olischl{\"a}ger, Hans and Habermann, Daniel and Kucharsk\'{y}, \v{S}imon and Elsem{\"u}ller, Lasse and Mishra, Aayush and Bracher, Niels and others},
  journal={arXiv preprint arXiv:2602.07098},
  year={2026}
}

@article{misra2019mish,
  title={Mish: A self regularized non-monotonic activation function},
  author={Misra, Diganta},
  journal={arXiv preprint arXiv:1908.08681},
  year={2019}
}

@article{zaheer2017deep,
  title={Deep sets},
  author={Zaheer, Manzil and Kottur, Satwik and Ravanbakhsh, Siamak and Poczos, Barnabas and Salakhutdinov, Russ R and Smola, Alexander J},
  journal={Advances in neural information processing systems},
  volume={30},
  year={2017}
}
}

%%%%%%%%%%%%%%%%%%%%%%%%%%%%%%%%%%%%%%%%%%%%%%%%%%%%%%%%%%%%

\appendix

\section{Conceptual comparison of the case studies}
\label{app:case-studies}

\begin{figure}
    \centering
    \includegraphics[width=\linewidth]{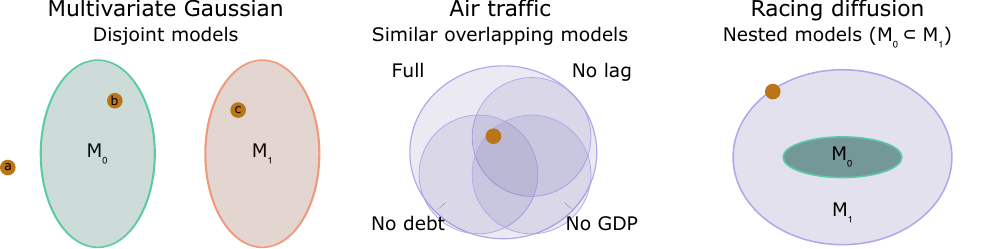}
    \caption{A conceptual comparison of the three case studies. Large colored circles show a simplified representation of the data distribution $p(y \mid M_k)$ implied by a specific model. Simulation-based losses are computed on data from these distributions. Small orange circles depict a representation of the location where the empirical (in the multivariate Gaussian case simulation test) data lie with respect to these predictive distributions.}
    \label{fig:case_studies_concept}
\end{figure}

The case studies in this article cover three common scenarios in comparing statistical models. \autoref{fig:case_studies_concept} illustrates the three case studies in terms of their relationship between the predictive distributions of the candidate models and the location of the evaluation data. 

\paragraph{Multivariate Gaussian: Disjoint models.} The two models have largely non-overlapping predictive distributions, as illustrated by the two separate ellipses. This separation creates a fundamental challenge for NPE and NPLE: When trained only with the simulation-based loss, any given data set is OOD for at least one of the two models. SC training therefore tremendously improves estimates in this scenario, especially in the neighborhood of the SC training data. For example, SC trained on data close to data point (a) will most likely reduce error for data point (a), but we also see benefits on data point (b) as the network representing $\mathcal{M}_1$ learns to extrapolate in the direction of $\mathcal{M}_0$. However, estimates for data point (c) remain inaccurate, as the network representing $\mathcal{M}_0$ learns to generalize towards (a) but not towards $\mathcal{M}_1$. For NPMP and NEE, the situation is different: the networks are trained on simulations from both $\mathcal{M}_0$ and $\mathcal{M}_1$, making data points (b) and (c) both in-distribution even without SC training. On the other hand, for NPMP and NEE, we observed that the networks' extrapolation bias for data point (a) is harder to correct with SC training.

\paragraph{Air traffic: Overlapping models.} The four models differ in which predictors are included (dropping one predictor at a time from the full model), resulting in predictive distributions that are similar and substantially overlapping, as shown by the four largely coincidental circles. The empirical data happens to fall close of the support of all four models. In this configuration, even NPE and NPLE produce accurate estimates with SC training, since the evaluation data is approximately in-distribution for every model; these approaches even outperform NPMP. SC training can provide improvements for NPE and NPLE, but the effect is quite small.

\paragraph{Racing diffusion model: Nested models.} The null model $\mathcal{M}_0$ is a special case of the alternative model $\mathcal{M}_1$, representing a common scenario of nested models. The orange dot represents empirical data for which the evidence in favor of the alternative model is so extreme that posterior model probabilities approach one. In this setting, NPMP correctly classifies data as coming from the alternative model even without SC training, while NPE, NPLE, and NEE can do so with appropriate SC training. Accurate estimation of Bayes factors is harder for all methods than just pointing at the better model, and only NPE shows clear improvements using SC training.

\section{Multivariate Gaussian}
\label{app:gaussian}

Neural posterior estimator, neural likelihood estimator, and neural evidence estimators were implemented as normalizing flows \citep{dinh2016density} with six affine coupling layers, each comprising of two dense layers with 256 hidden units and mish activations \citep{misra2019mish}. NPMP was implemented as an MLP with two dense layers of 256 units each and mish activations and residual connections, and an output layer with 2 units and softmax activation.

In total, we trained each network for 50 epochs, 32 steps per epoch. We used online simulation-based training with batch size of 64 per model (NPE and NPLE trained with batch size 64, NEE and NPMP with batch size 128, of which about 50\% were drawn from $\mathcal{M}_0$ and 50\% from $\mathcal{M}_1$). For the self-consistency loss, we used 16 pre-simulated datasets from each model that was included in the SC set (e.g., \emph{SC(-6)} was trained on 16 simulations from model with $\mu_\mu =- 6$, \emph{SC(-2, +2} was trained on 16 simulations with $\mu_\mu=-2$ and 16 simulations from $\mu_\mu=+2$, and so forth). Because the SC loss can be unstable when networks are initialized randomly, we warm-start its weight: $\lambda_{\mathrm{SC}}=0$ during the first epoch, increase it linearly from 0 to 1 during the second epoch, and keep it fixed at 1 thereafter. We use $S=16$ posterior samples to approximate the variance of the log marginal likelihood in the SC loss (Equation~\ref{eq:sc_loss}), and $S=128$ samples at test time to approximate the log marginal likelihood.

For NPLE+, the likelihood network was pretrained using a model with uniform prior on $\mu_\mu$ but with the same likelihood. The likelihood network was trained online for 50 epochs, 32 steps per epoch, batch size 64, and subsequently frozen. The rest of the training (of the posterior network) followed the same procedure as methods mentioned above.

Training all models in all conditions took about 2h on Apple M1 Pro personal laptop, without using GPU acceleration. Inference time of the ABI methods was negligible compared to training. The neural estimators were implemented using the Python library \texttt{BayesFlow} \citep{bayesflow_2020_original,bayesflow_2023_software,kuhmichel2026bayesflow}.

\begin{figure}
    \centering
    \includegraphics[width=\linewidth]{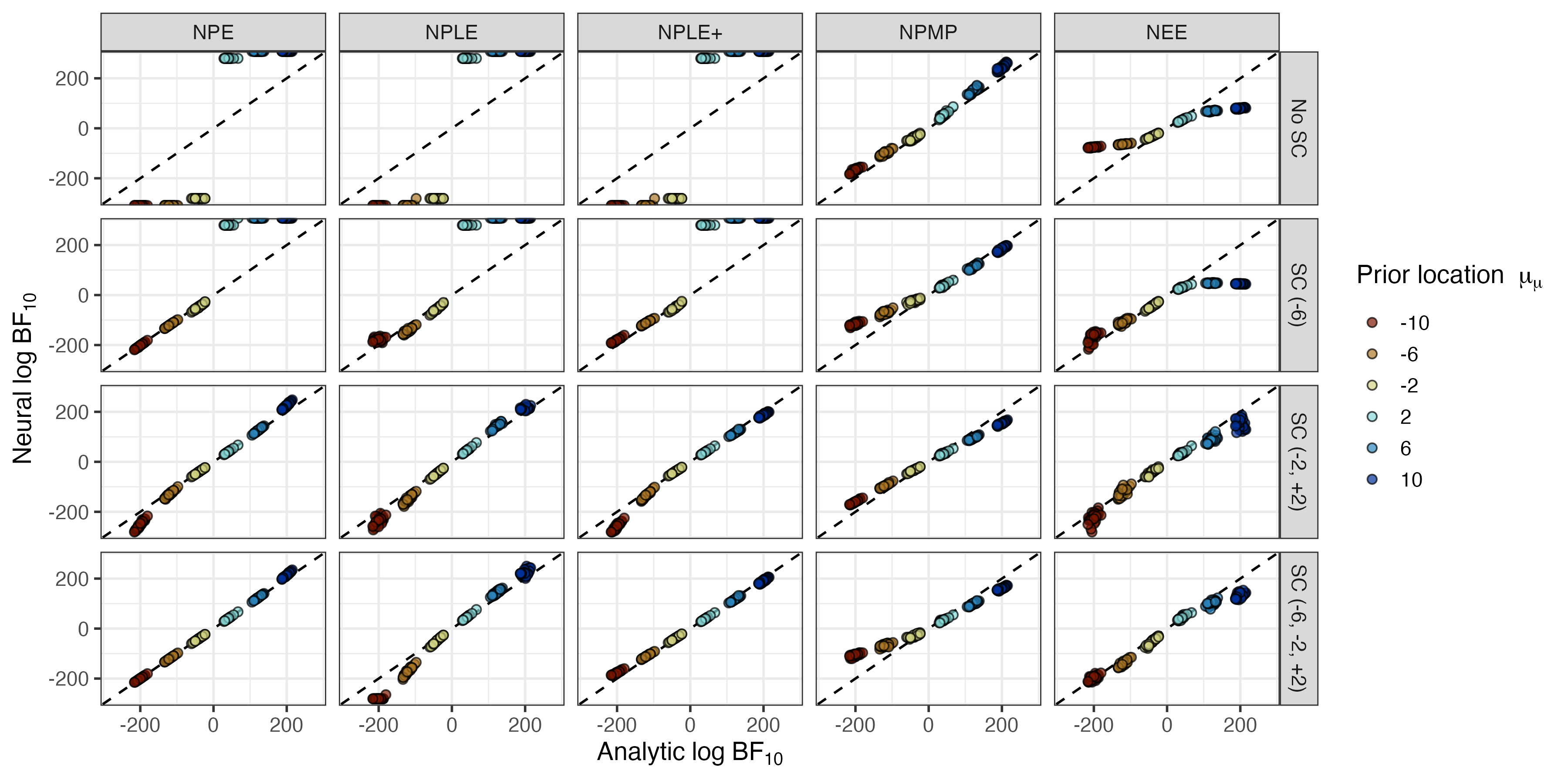}
    \caption{\textbf{Gaussian case study.} Neural estimates of $\log \text{BF}_{10}$ compared to its analytic form, evaluated on datasets with different prior locations. Data along the $y$-axis were squished at -300 and 300.}
    \label{fig:gaussian_log_bf}
\end{figure}

\paragraph{Additional results.} \autoref{fig:gaussian_log_bf} shows the approximated Bayes factors against the analytic Bayes factors, and provides an alternative view of results shown in \autoref{fig:gaussian_log_bf_diff} which shows error of the neural estimates.

\begin{figure}
    \centering
    \includegraphics[width=\linewidth]{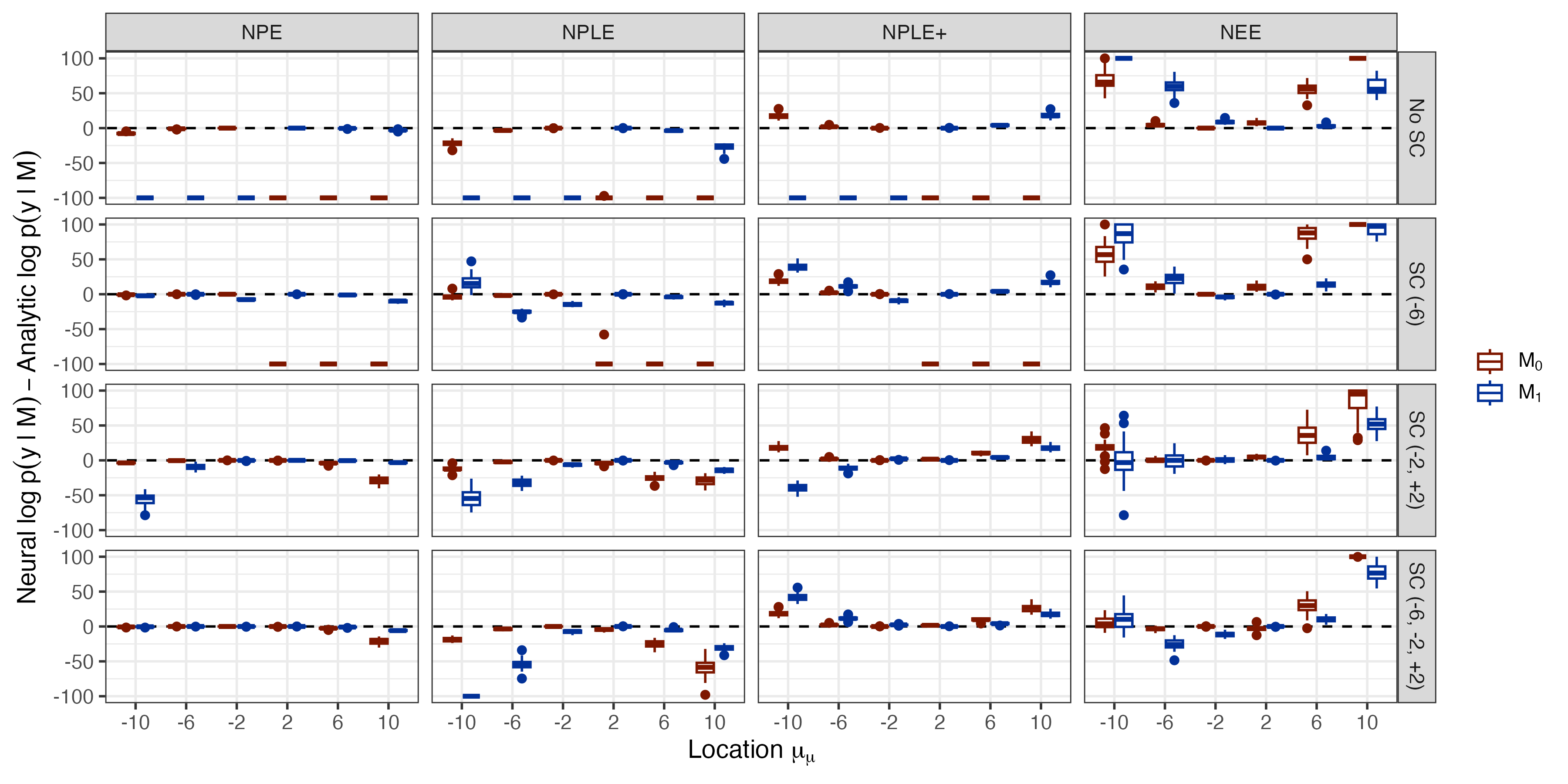}
    \caption{\textbf{Gaussian case study.} Error of the neural approximate log marginal likelihoods for the two models, evaluated on datasets with different prior locations. Data along the $y$-axis were squished at -100 and 100.}
    \label{fig:gaussian_log_ml_diff}
\end{figure}

\autoref{fig:gaussian_log_ml_diff} shows the error of the log marginal likelihood estimates, and provide more detailed picture of the main results. For instance, in the case of NPE without SC training, either the posterior network trained on simulations from $\mathcal{M}_0$ or from $\mathcal{M}_1$ has a high error, which leads to the Bayes factor estimates being incorrect for any data set, even though inferences for the two individual models are reasonably accurate locally for a subset of the data space.

\section{Air passenger traffic forecasting}
\label{app:air_traffic}

We fitted a set of first-order autoregressive models of varying complexity:
\begin{subequations}
\begin{align}
M_1 (\text{Full}): \quad y_{j, t+1} &\sim \mathcal{N}\!(\alpha_j + y_{j, t}\beta_j + u_{j, t}\gamma_j + w_{j,t}\delta_j, \sigma_j),  \label{eq:full} \\
M_2 (\text{No lag}): \quad y_{j, t+1} &\sim \mathcal{N}\!(\alpha_j + u_{j, t}\gamma_j + w_{j,t}\delta_j, \sigma_j), \label{eq:no_y} \\
M_3 (\text{No debt}): \quad y_{j, t+1} &\sim \mathcal{N}\!(\alpha_j + y_{j, t}\beta_j + w_{j,t}\delta_j, \sigma_j), \label{eq:no_u} \\
M_4 (\text{No GDP}): \quad y_{j, t+1} &\sim \mathcal{N}\!(\alpha_j + y_{j, t}\beta_j + u_{j, t}\gamma_j, \sigma_j), \label{eq:no_w} 
\end{align}
\end{subequations}
where $y_{j,t+1}$ denotes the year-to-year change in air passenger traffic for country $j$. The covariate $u_{j,t}$ corresponds to household debt (as a percentage of GDP), and $w_{j,t}$ represents real GDP per capita. Model~\ref{eq:full} includes all predictors, while Models~\ref{eq:no_u}--\ref{eq:no_y} remove one predictor each, allowing us to assess the marginal contribution of lagged traffic, household debt, and GDP per capita.

For these models we use independent priors following \citet{mishra_robust_2025}:
\begin{align*}
    \alpha_j &\sim \mathcal{N}\!(0, 0.5) \hspace{1.5cm} \beta_j \sim \mathcal{N}\!(0, 0.2) \\ 
    \gamma_j &\sim \mathcal{N}\!(0, 0.5) \hspace{1.5cm} \delta_j \sim \mathcal{N}\!(0, 0.5)\\
    \mathrm{log}(\sigma_j) &\sim \mathcal{N}\!(-1, 0.5)   
\end{align*}
The parameters $\alpha_j$ capture country-specific intercepts, $\beta_j$ represent autoregressive coefficients, $\gamma_j$ and $\delta_j$ quantify the coefficients of household debt and GDP per capita, and $\sigma_j$ represents residual variation.

Both neural posterior, neural likelihood, and neural evidence estimators were implemented as normalizing flows \citep{dinh2016density}. Posterior and likelihood were implemented with six affine coupling layers, and evidence network with eight coupling layers. Each coupling layer contains two dense layers with 256 hidden units and mish activations \citep{misra2019mish}. The classification network for NPMP was implemented as an MLP with 2 dense layers of 256 units each and mish activations and residual connections, and an output layer with 2 units and softmax activation. The data are coming from an AR(1) process but we included the lagged values of the dependent variables as one of the predictors, effectively making the individual observations exchangeable. The likelihood network was trained to predict the likelihood per-observation and the total likelihood was calculated as a sum of the individual log-likelihoods. The summary network follows a DeepSet architecture \citep{zaheer2017deep} comprising two equivariant modules (each with two dense layers and 64 hidden units), an invariant module with two dense layers and 32 hidden units in both the inner and outer components, an additional invariant module with two dense layers of 32 hidden units, and an output layer with 30 units. All layers except the output layer use SiLU activations.

In total, we trained each network for 30 epochs, 500 steps per epoch. We used online simulation-based training with batch size of 32 per model (NPE and NPLE trained with batch size 32, NEE and NPMP with batch size 128, of which about a quarter of the simulations were from each of the compared models). For the self-consistency loss, we pre-simulated 20 data sets from each of the four models, and have 15 empirical data sets (from the 15 countries). Table~\ref{tab:air_traffic_sc_conditions} shows conditions of SC training, depending on what subsets of the three data sources were used for computing the SC loss. Note that the data sets that ended up used for SC training are labeled as ``Train'' and data sets that were not used for SC training were labeled as ``Test''. Because the SC loss can be unstable when networks are initialized randomly, we warm-start its weight: $\lambda_{\mathrm{SC}}=0$ for epochs 1–10, increase it linearly from 0 to 1 over epochs 11–20, and keep it fixed at 1 thereafter. We use $S=16$ posterior samples to approximate the variance of the log marginal likelihood in the SC loss (Equation~\ref{eq:sc_loss}), and $S=128$ samples at test time to approximate the log marginal likelihood.

Training all models in all conditions took about 24h on Apple M1 Pro personal laptop, without using GPU acceleration. Fitting the models with MCMC and estimating marginal likelihoods with bridge sampling for gold standard comparison took about 1h on the same system. Inference time of the ABI methods was negligible compared to training and fitting with MCMC. The neural estimators were implemented using the Python library \texttt{BayesFlow} \citep{bayesflow_2020_original,bayesflow_2023_software,kuhmichel2026bayesflow}.

\begin{table*}[t]
    \caption{\textbf{Air traffic case study.} Self-consistency training conditions: the number of datasets from different sources used for SC training in different conditions. Highlighted in bold are conditions that are discussed in the main text, the rest of the results are discussed only in the appendix.}
    \centering
    \begin{tabular}{lrrrrr}
    \toprule
       SC Condition  &  Full model & No lag & No debt & no GDP & Empirical  \\
    \midrule
       \textbf{None}  & 0 & 0 & 0 & 0 & 0  \\
       \textbf{Simulated} & 4 & 4 & 4 & 4 & 0  \\
       \textbf{Empirical (8)} & 0 & 0 & 0 & 0 & 8  \\
       Empirical (15) & 0 & 0 & 0 & 0 & 15  \\
       \textbf{Simulated + Empirical (8)} & 4 & 4 & 4 & 4 & 8  \\
       Simulated + Empirical (15) & 4 & 4 & 4 & 4 & 15  \\
    \bottomrule
    \end{tabular}
    \label{tab:air_traffic_sc_conditions}
\end{table*}

\begin{figure}
    \centering
    \includegraphics[width=\linewidth]{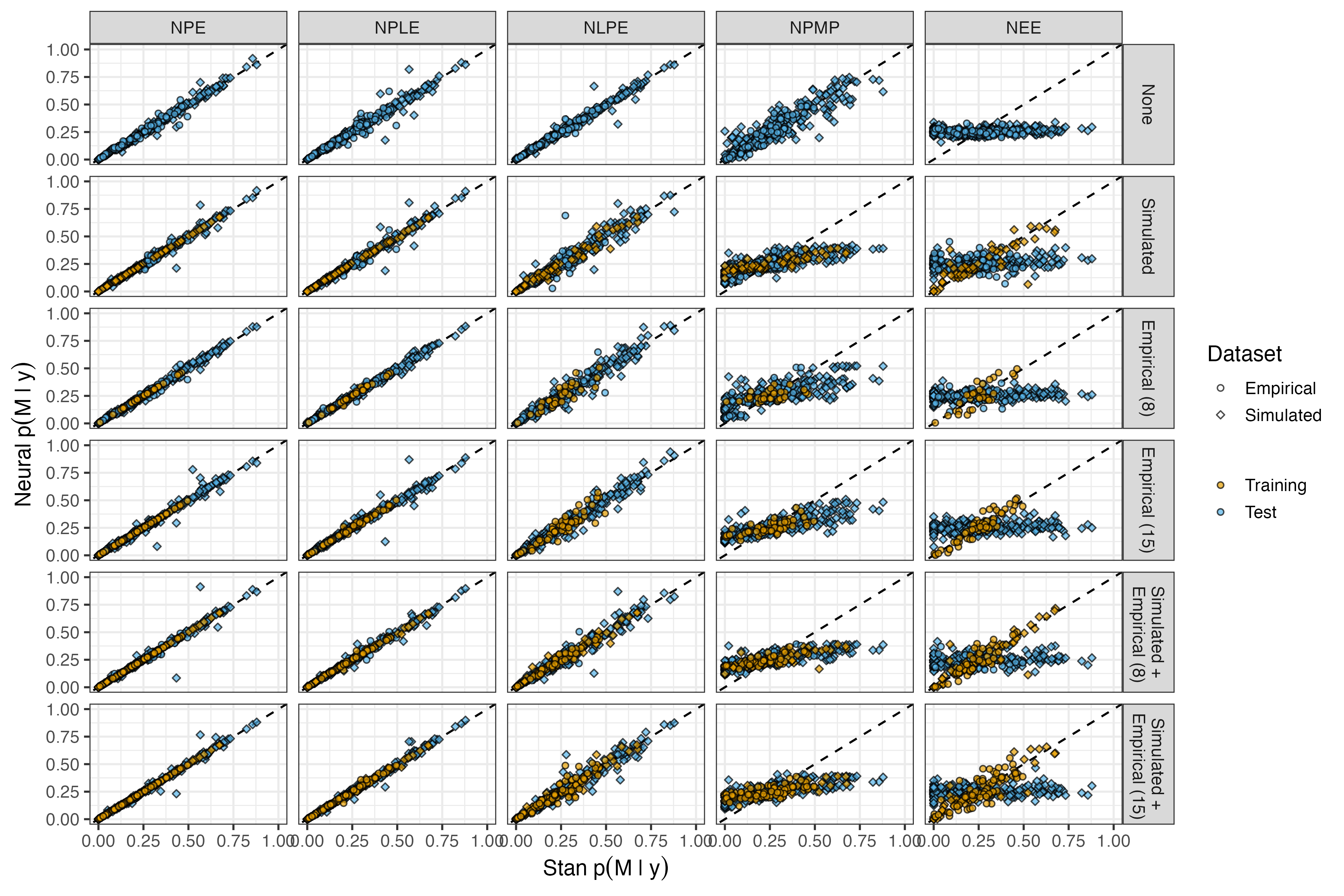}
    \caption{\textbf{Air traffic case study.} Posterior model probability estimates of the four methods, compared to the gold standard, evaluated on empirical data and simulations from the four models, including the additional method \emph{NLPE} and additional SC training conditions.}
    \label{fig:air_traffic_pmp}
\end{figure}

\begin{figure}
    \centering
    \includegraphics[width=\linewidth]{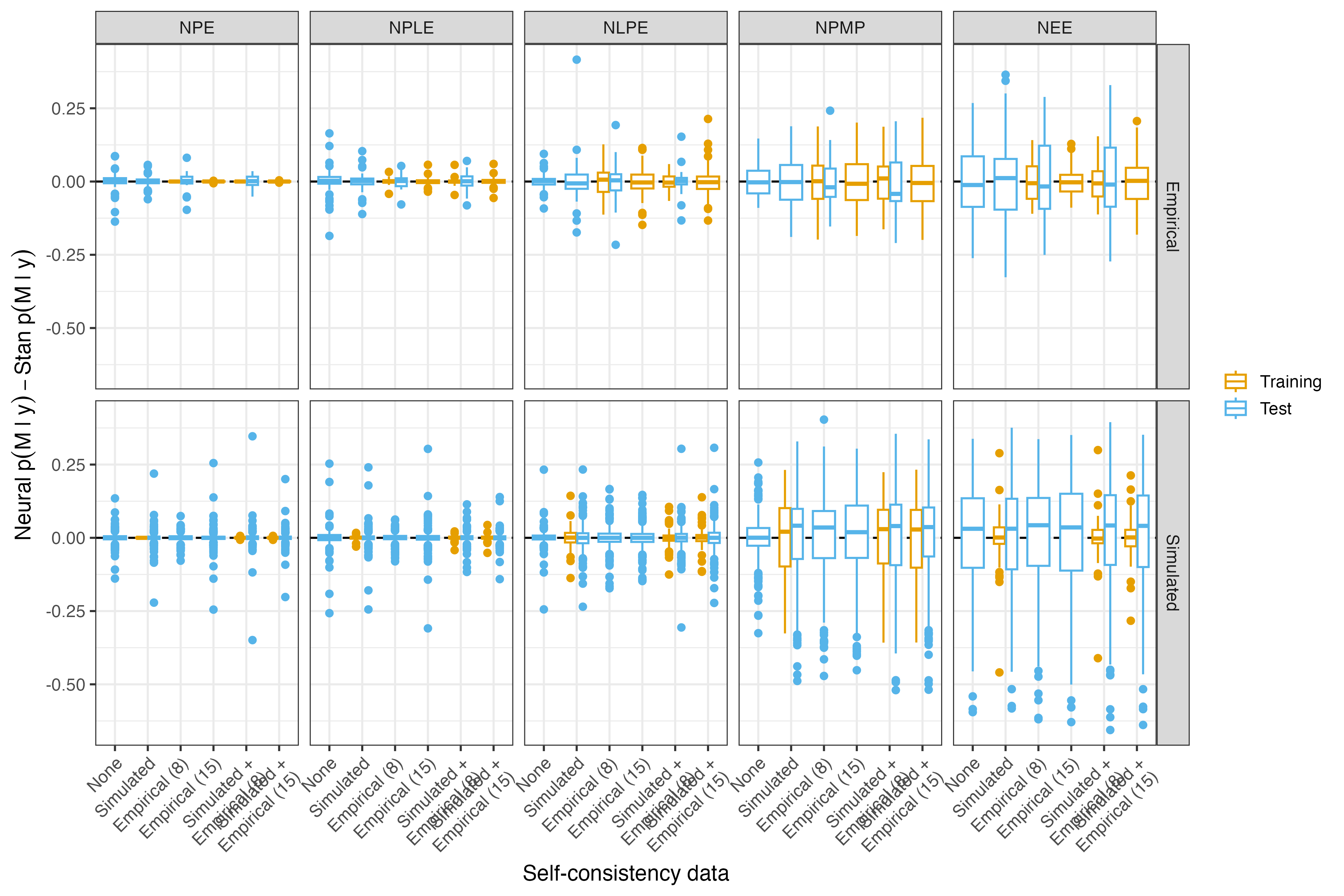}
    \caption{\textbf{Air traffic case study.} The error of the approximate posterior model probabilities of the four models including the additional method \emph{NLPE} and additional SC training conditions.}
    \label{fig:air_traffic_pmp_diff}
\end{figure}

\paragraph{Additional results.} We considered an alternative method for NPLE where the likelihood network instead of the posterior network is trained with the SC loss. This method is labeled \textbf{NLPE}. The results of this method are compared on \autoref{fig:air_traffic_pmp} and \ref{fig:air_traffic_pmp_diff}. These figures also show results from additional SC training conditions. Overall, NPLE performs better than NLPE, which suggests that training the posterior network with SC has more benefits than training the likelihood network with SC. 

\section{Racing Diffusion Model of Decision Making}
\label{app:diffusion}

The racing diffusion model \citep{tillman2020sequential} assumes that each response option (here, correct vs.\ incorrect) is associated with a dedicated noisy evidence accumulator. Each accumulator begins at zero and evolves according to a Wiener process with drift (where $W_t$ is the standard Wiener process),
\begin{equation}
X_t = \nu~\mathrm{d}t + \sigma~\mathrm{d}W_t,
\qquad X_0 = 0.
\end{equation}
A response is initiated when one of the accumulators first reaches the decision boundary $\alpha$. The observed response time is the hitting time of the winning accumulator plus a non-decision component $t_0$. Thus, the model produces response times ($rt$, in seconds) and accuracies ($d$, correct/incorrect) on each trial. For training purposes, we recoded incorrect responses as negative values of the corresponding $rt$, enabling the networks to condition on a single variable capturing both speed and accuracy.

The experiment alternated between speed and accuracy instruction blocks. Within each block, participants were instructed either to respond as quickly as possible or to prioritize accuracy. Each block comprised 96 trials. For the present analysis, we selected the first four blocks (two per condition) for each participant, yielding datasets of 384 trials (192 per condition).

Under the null model, $\alpha$ is held constant across conditions:
\begin{equation}
\begin{aligned}
M_0: \text{Fixed }\alpha \text{ across conditions}, \
\log \alpha \sim \mathcal{N}(0, 0.5),
\end{aligned}
\end{equation}
whereas under the alternative model, the decision boundary varies by condition:
\begin{equation}
\begin{aligned}
M_1: \text{Varying }\alpha \text{ across conditions}, \
\log \alpha_{\text{speed}} &\sim \mathcal{N}(0, 0.5), \
\log \alpha_{\text{accuracy}} &\sim \mathcal{N}(0, 0.5).
\end{aligned}
\end{equation}
The alternative model thus reflects the hypothesis that participants adapt their response caution (i.e., the amount of evidence required before responding) in accordance with task instructions.

Under both models, the nuisance parameters follow the priors
\begin{equation}
\begin{aligned}
\log \nu_{\text{correct}} &\sim \mathcal{N}(0, 0.5), \
\log \nu_{\text{incorrect}} &\sim \mathcal{N}(0, 0.5), \
\mathrm{logit},\tau &\sim \mathcal{N}(0, 1),
\end{aligned}
\end{equation}
with $\sigma = 1$. The parameter $\tau$ determines the non-decision time via
\begin{equation}
t_0 = \min(rt)\frac{\tau}{1-\tau}.
\end{equation}
This reparameterization permits estimation of $\mathrm{logit}~\tau$ on an unconstrained scale while ensuring $t_0 \in (0, \min(rt))$.

Both neural posterior, neural likelihood, and neural evidence estimators were implemented as normalizing flows \citep{dinh2016density} with six affine coupling layers. Each coupling layer contains two dense layers with 256 hidden units and mish activations \citep{misra2019mish}. The classification network for NPMP was implemented as an MLP with 2 dense layers of 256 units each and mish activations and residual connections, and an output layer with 2 units and softmax activation. The summary network follows a DeepSet architecture \citep{zaheer2017deep} comprising two equivariant modules (each with two dense layers and 64 hidden units), an invariant module with two dense layers and 32 hidden units in both the inner and outer components, an additional invariant module with two dense layers of 32 hidden units, and an output layer with 30 units. All layers except the output layer use SiLU activations.

In total, we trained each network for 50 epochs, 64 steps per epoch. We used online simulation-based training with batch size of 64 per model (NPE and NPLE trained with batch size 64, NEE and NPMP with batch size 128, of which about 50\% were drawn from the null and 50\% from the alternative models, respectively). For the self-consistency loss, we pre-simulated 17 data sets from the null, 17 data sets from the alternative, and have 17 empirical data sets. Table~\ref{tab:diffusion_sc_conditions} shows conditions of SC training, depending on what subsets of the three data sources were used for computing the SC loss. Note that the data sets that ended up used for SC training are labeled as ``Training'' and data sets that were not used for SC training were labeled as ``Test''. Because the SC loss can be unstable when networks are initialized randomly, we warm-start its weight: $\lambda_{\mathrm{SC}}=0$ for epochs 1–20, increase it linearly from 0 to 1 over epochs 21–30, and keep it fixed at 1 thereafter. We use $S=16$ posterior samples to approximate the variance of the log marginal likelihood in the SC loss (Equation~\ref{eq:sc_loss}), and $S=128$ samples at test time to approximate the log marginal likelihood.

Training all models in all conditions took about 32h on Apple M1 Pro laptop, without using GPU acceleration. Fitting the models with MCMC and estimating marginal likelihoods with bridge sampling for gold standard comparison took about 6h on the same system. Inference time of the ABI methods was negligible compared to training and fitting with MCMC. The neural estimators were implemented using the Python library \texttt{BayesFlow} \citep{bayesflow_2020_original,bayesflow_2023_software,kuhmichel2026bayesflow}.

\begin{table*}[t]
    \caption{\textbf{Racing diffusion model case study.} Self-consistency training conditions: the number of datasets from different sources used for SC training in different conditions. Highlighted in bold are conditions that are discussed in the main text, the rest of the results are discussed only in the appendix.}
    \centering
    \begin{tabular}{lrrr}
    \toprule
       SC Condition  &  Null model & Alternative model & Empirical  \\
    \midrule
       \textbf{None}  &  0 & 0 & 0  \\
       \textbf{Simulated (10)} &  10 & 10 & 0  \\
       Simulated (17) & 17 & 17 & 0  \\
       \textbf{Empirical (10)} & 0 & 0 & 10  \\
       Empirical (17) & 0 & 0 & 17  \\
       \textbf{Simulated + Empirical (10)} &  10 & 10 & 10  \\
       Simulated + Empirical (17) & 17 & 17 & 17  \\
    \bottomrule
    \end{tabular}
    \label{tab:diffusion_sc_conditions}
\end{table*}

\begin{figure}
    \centering
    \includegraphics[width=\linewidth]{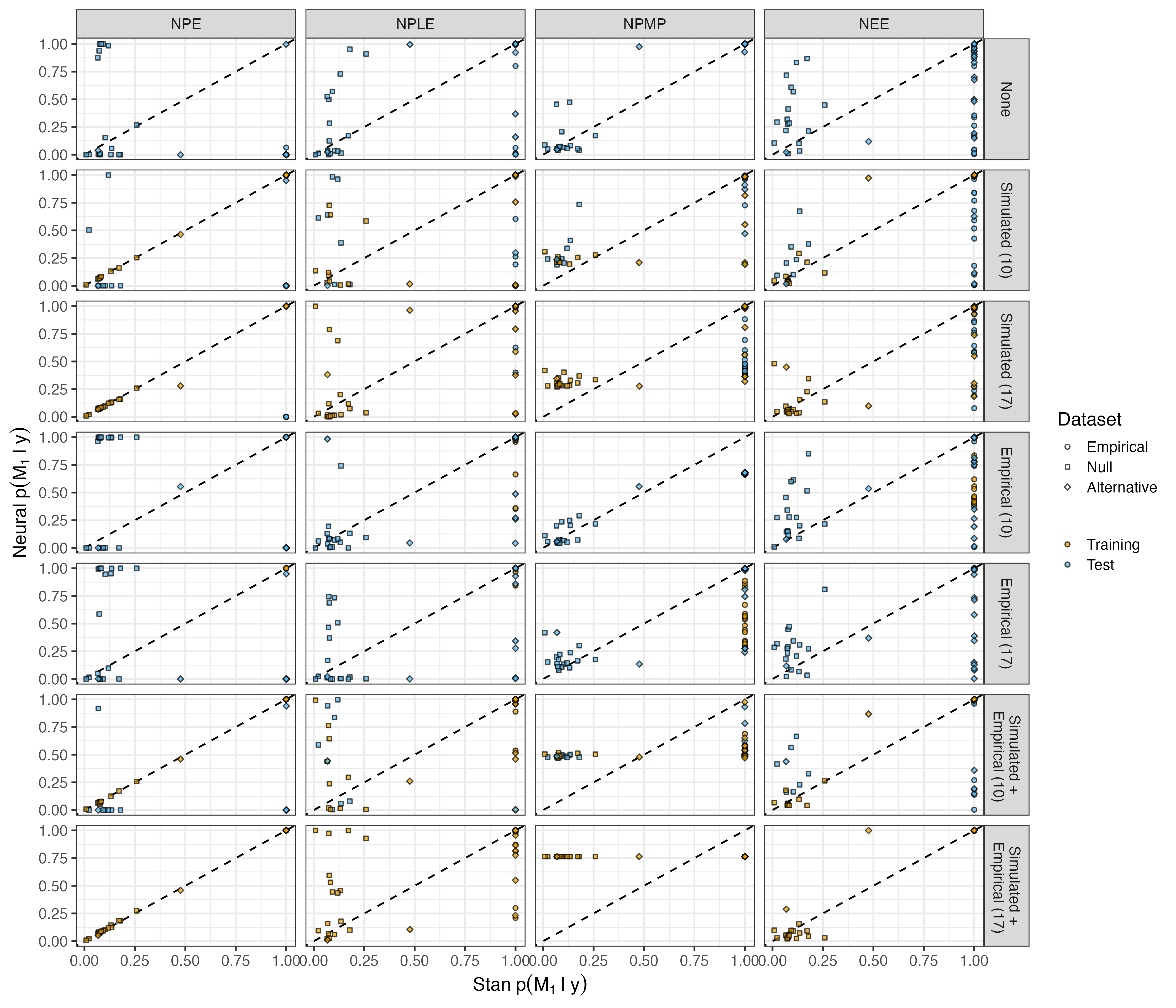}
    \caption{\textbf{Racing diffusion case study.}  The posterior model probability of the alternative model compared against the gold standard, including additional SC training conditions.}
    \label{fig:diffusion_pmp}
\end{figure}

\begin{figure}
    \centering
    \includegraphics[width=\linewidth]{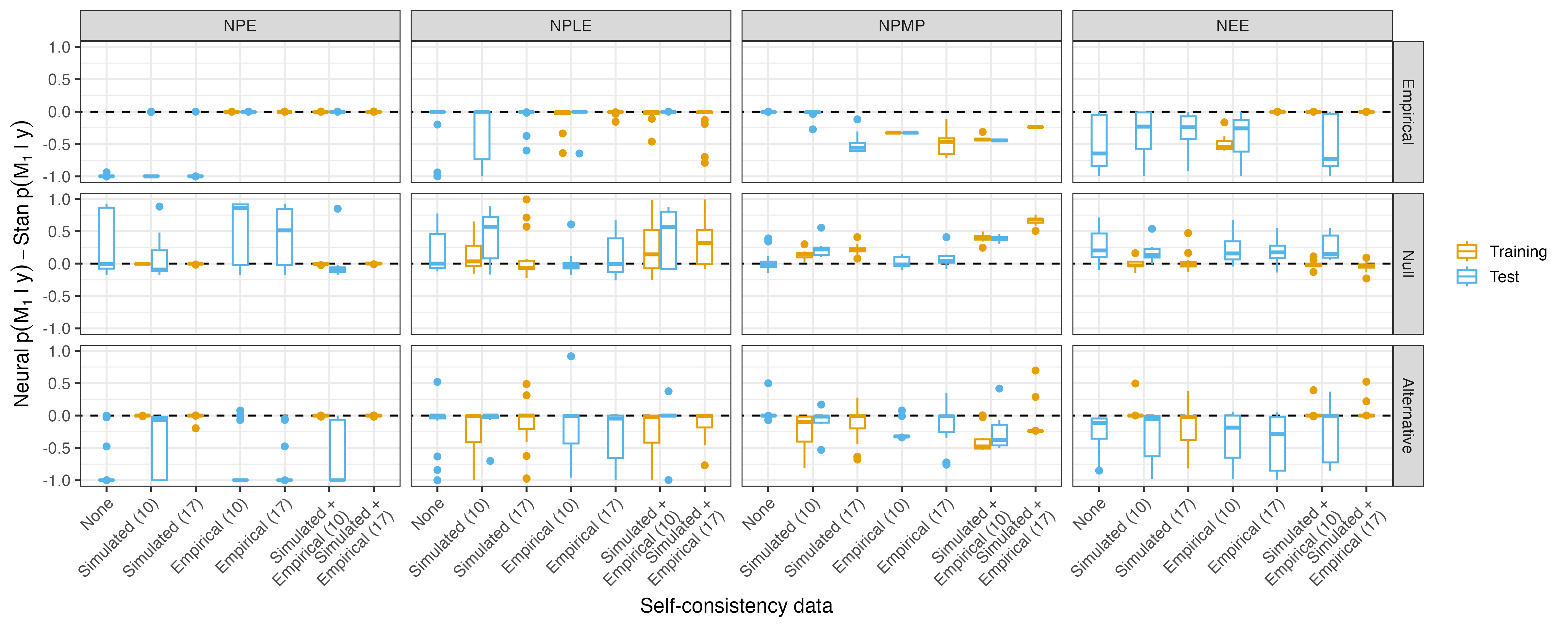}
    \caption{\textbf{Racing diffusion case study.} The error of the posterior model probability of the alternative model estimates, including additional SC training conditions.}
    \label{fig:diffusion_pmp_diff}
\end{figure}

\paragraph{Additional results.} \autoref{fig:diffusion_pmp} and \ref{fig:diffusion_pmp_diff} show the results on the posterior model probability scale. For the empirical data, the evidence is so decisive (in favour of the alternative model), that the posterior model probability is approaches 1. While NPMP estimates this accurately, NPE, NPLE, and NEE networks do so only with some form of SC training. Similar pattern can be found when the data is generated from the alternative model. When the data is generated from the null model, NPE estimates are the most accurate, provided that data from the null model are included in the SC training set. NPLE struggles to estimate the posterior model probabilities accurately even with SC training. NPMP provides somewhat acceptable results without SC, but with relatively high variance. NEE fails to estimate posterior model probabilities correctly similar to NPLE, SC training nudges it towards estimates similar to NPLE.

\section{Additional Gaussian experiment}
\label{app:gaussian_addition}

Here we include additional results from previous experimentation with NPE, NPLE and NPMP. In this example, we are in a scenario where the two models are nested, and are always misspecified to the SC training (and evaluation) data. We vary the dimensionality of the data to see how results of SC training translate with increasing dimensions.

Two random vectors are drawn from a unit variance multivariate Gaussian,
\begin{equation}
\begin{aligned}
    \mathbf{x}_i & \sim \mathcal{N}(\boldsymbol\mu_x,\, \mathbf{I}_D) & \\
    \mathbf{y}_i & \sim \mathcal{N}(\boldsymbol\mu_y,\, \mathbf{I}_D) &\text{for }i=1,\dots,N.
\end{aligned}
\end{equation}
Under the null model, the means of the two vectors are set to be equal, $M_0: \boldsymbol{\mu_x} = \boldsymbol{\mu_y} \sim \mathcal{N}(\mathbf{0},\, \sigma_\mu^2 \mathbf{I}_D)$. Under the alternative model, the means are drawn independently, $M_1: \boldsymbol{\mu_x} \sim \mathcal{N}(\mathbf{0},\, \sigma_\mu^2 \mathbf{I}_D),\ \boldsymbol{\mu_y} \sim \mathcal{N}(\mathbf{0},\, \sigma_\mu^2 \mathbf{I}_D)$. We vary $D \in \{1, 5, 10\}$ and $N = \{1, 10, 20\}$.

We train NPE, NPLE, NPMP with or without SC training. For the SC loss, we use $M=32$ datasets drawn from $M_1$, but the two random vectors shifted $\mathbf{x}' = \mathbf{x} + \delta$, $\mathbf{y}' = \mathbf{y} - \mathbf{\delta}$, with $\delta=3$. Both of the models are therefore misspecified, though in general, the data favors $M_1$.

During training, we either fix the prior scale ($\sigma_\mu^2 = 1$) or randomize it by sampling $\log \sigma_\mu^2 \sim \mathrm{Uniform}(-3,3)$. In the latter case, $\sigma_\mu$ is provided to the networks as an additional conditioning variable, effectively specifying different models with varying prior concentration. This variation alters the implied posterior while leaving the likelihood unchanged. By exposing the likelihood network in NPLE to data generated under a broader range of prior scales than $\sigma_\mu^2 = 1$, we expect the learned likelihood to become more robust to prior misspecification.

\begin{figure}
    \centering
    \includegraphics[width=\linewidth]{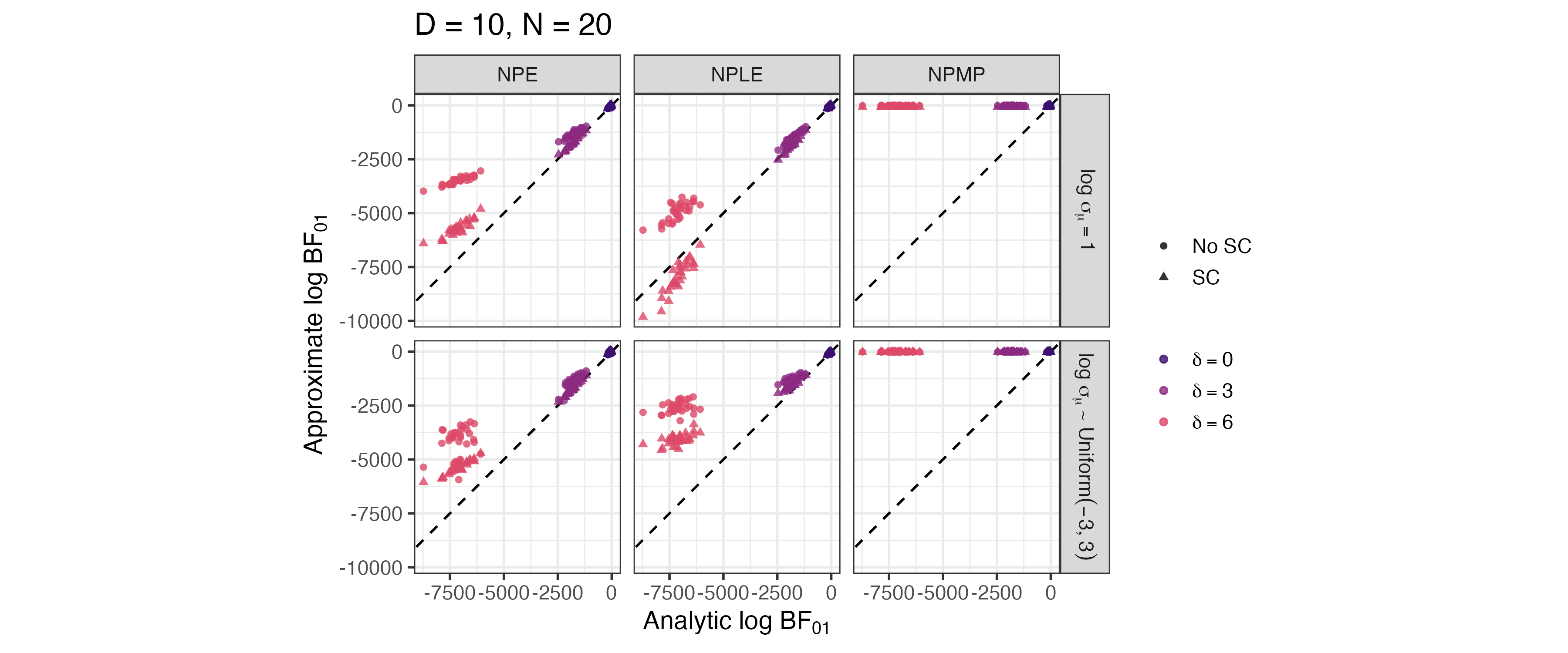}
    \caption{\textbf{Additional Gaussian experiment.} Approximate log BF against the analytic log BF for NPE, NPLE, NPMP, trained on narrow ($\sigma_\mu=1$) or wide ($\log\sigma_\mu\sim\text{Uniform(-3, 3}$) priors, evaluated on data generated with $\delta=0$ (in-distribution for $M_1$), $\delta=3$ (in-distribution for SC training), or $\delta=6$ (out of distribution).}
    \label{fig:gaussian_additional_example}
\end{figure}

\paragraph{Results.} Figure~\ref{fig:gaussian_additional_example} shows the results for an experiment with $D=10$, $N=20$. When the test data is generated from $M_1$ ($\delta=0$), the approximated log Bayes factors are relatively precise for all methods. Once the models are misspecified, NPMP is not able to approximate the log Bayes factor accurately, with or without SC training. SC training helps estimating log Bayes factors for NPE and NPLE, mainly when the test data is similar to that used for SC training ($\delta=3$). When the test data is even more extreme ($\delta=6$), SC improves the estimates, although its bias cannot be completely rectified. Training with a wider range of priors ($\log \sigma_\mu^2 \sim \mathrm{Uniform}(-3,3)$) tends to have relatively negligible effects on improving estimates.

\begin{figure}
    \centering
    \includegraphics[width=\linewidth]{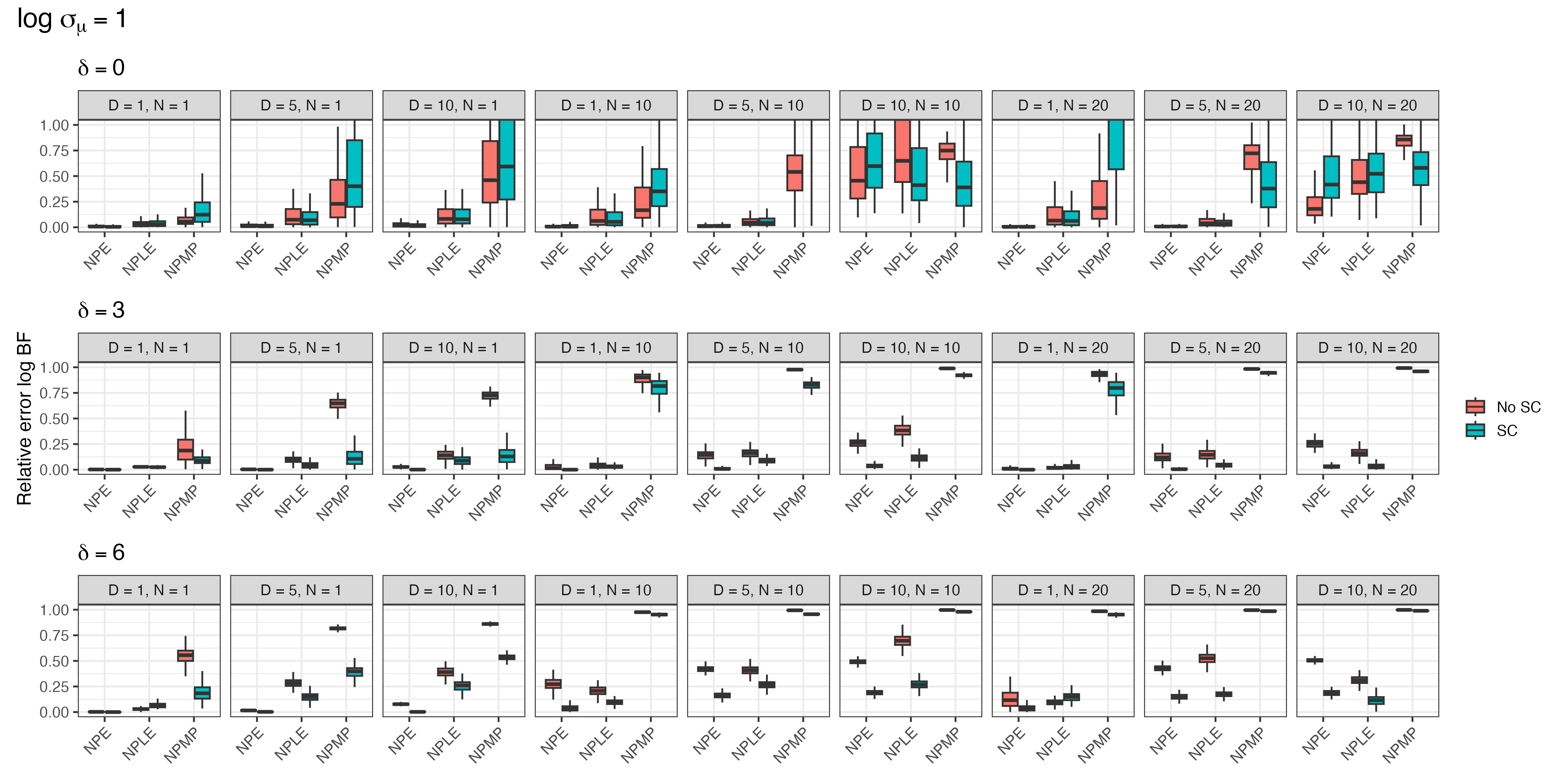}
    \caption{\textbf{Additional Gaussian experiment.} Relative error of the neural approximate log BF summarized across varying dimensionality of the data.}
    \label{fig:gaussian_additional_narrow}
\end{figure}

\begin{figure}
    \centering
    \includegraphics[width=\linewidth]{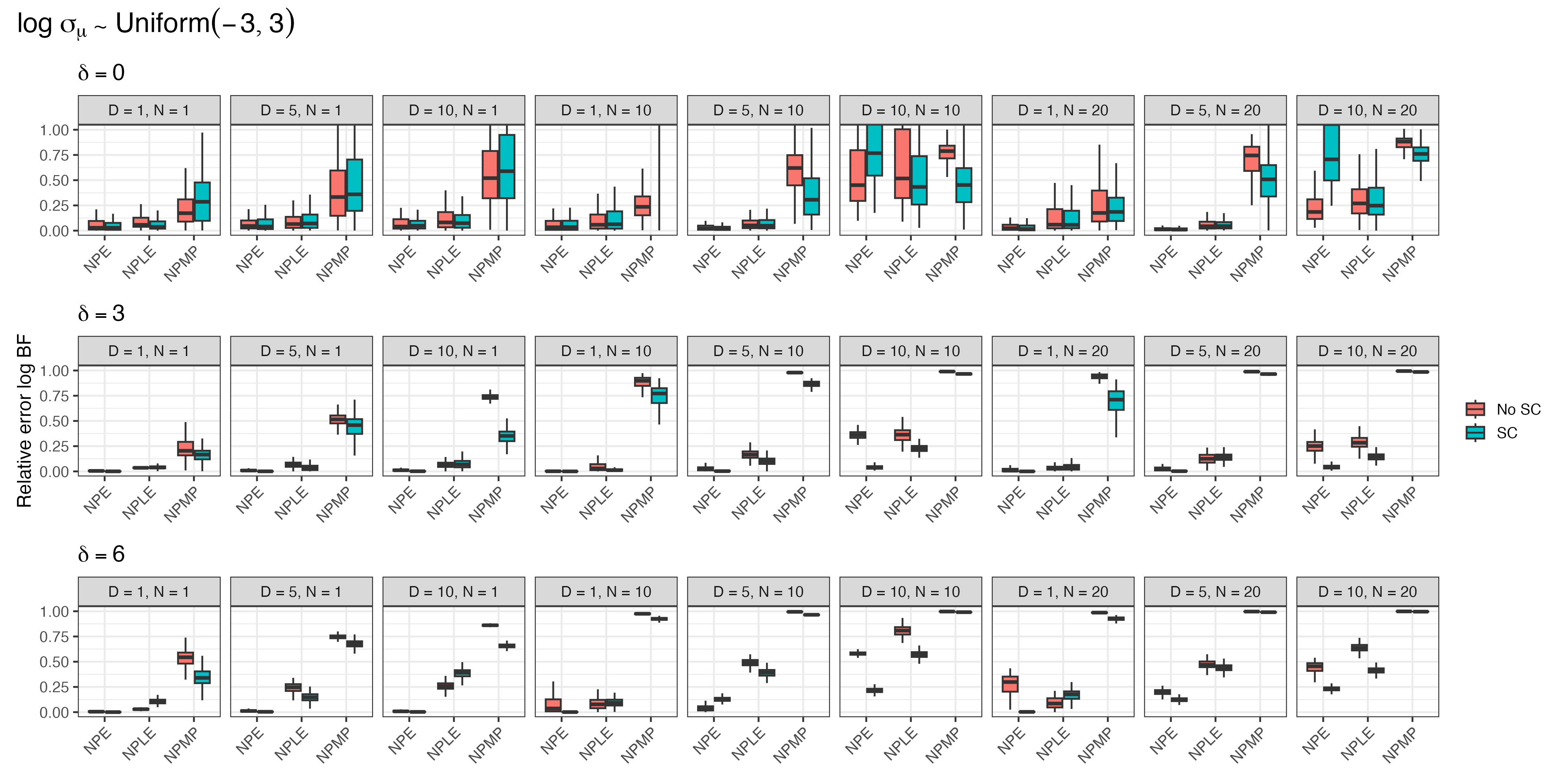}
    \caption{\textbf{Additional Gaussian experiment.} Relative error of the neural approximate log BF summarized across varying dimensionality of the data. Increasing the width of the prior has small impact on improving results of SC training.}
    \label{fig:gaussian_additional_wide}
\end{figure}

Simulations with varying dimensionality of the data are summarised on Figures~\ref{fig:gaussian_additional_narrow} and \ref{fig:gaussian_additional_wide}. Overall, NPE and NPLE vastly outperform NPMP in the open-world setting. SC improves log Bayes factor approximation especially for NPE and on test data similar to that used for SC training ($\delta=3$). NPLE tends to benefit from SC as well, though the effect is less consistent across different simulations.

%%%%%%%%%%%%%%%%%%%%%%%%%%%%%%%%%%%%%%%%%%%%%%%%%%%%%%%%%%%%

% \newpage
% \input{neurips_2026/checklist}

\end{document}